\documentclass[lettersize,journal]{IEEEtran}
\usepackage{amsmath,amsfonts}
\usepackage{algorithmic}
\usepackage{algorithm}
\usepackage{array}
\usepackage[caption=false,font=normalsize,labelfont=sf,textfont=sf]{subfig}
\usepackage{textcomp}
\usepackage{stfloats}
\usepackage{url}
\usepackage{verbatim}
\usepackage{graphicx}
\usepackage{cite}
\usepackage{microtype}      
\usepackage{color}
\usepackage{pifont}
\usepackage[dvipsnames]{xcolor}

\begin{document}

\title{CLIP the Divergence: Language-guided Unsupervised Domain Adaptation}

\author{Anonymous Submission}
\author{Jinjing Zhu, Yucheng Chen, Lin Wang
}

\markboth{Journal of \LaTeX\ Class Files,~Vol.~14, No.~8, August~2021}%
{Shell \MakeLowercase{\textit{et al.}}: A Sample Article Using IEEEtran.cls for IEEE Journals}


\maketitle

\begin{abstract}
Unsupervised domain adaption (UDA) has emerged as a popular solution to tackle the divergence between the labeled source and unlabeled target domains. Recently, some research efforts have been made to leverage large vision-language models, such as CLIP, and then fine-tune or learn prompts from them for addressing the challenging UDA task. 
In this work, we shift the gear to a new direction by directly leveraging CLIP to measure the domain divergence and propose a novel language-guided approach for UDA, dubbed as \textbf{CLIP-Div}. Our key idea is to harness CLIP to 1) measure the domain divergence via the acquired domain-agnostic distribution and 2) calibrate the target pseudo labels with language guidance, to effectively reduce the domain gap and improve the UDA model's generalization capability. Specifically, our major technical contribution lies in the proposed two novel language-guided domain divergence measurement losses: \textbf{absolute divergence} and \textbf{relative divergence}. These loss terms furnish precise guidelines for aligning the distributions of the source and target domains with the domain-agnostic distribution derived from CLIP. Additionally, we propose a \textbf{language-guided pseudo-labeling} strategy for calibrating the target pseudo labels. Buttressed by it, we show that a further implementation for self-training can enhance the UDA model's generalization capability on the target domain. CLIP-Div surpasses state-of-the-art CNN-based methods by a substantial margin, achieving a performance boost of \textbf{+10.3}\% on Office-Home, \textbf{+1.5}\% on Office-31, \textbf{+0.2}\% on VisDA-2017, and +\textbf{24.3}\% on DomainNet, respectively.  
\end{abstract}


\section{Introduction}
\IEEEPARstart{I}{n} recent years, significant advancements have been observed in various computer vision tasks by harnessing the power of deep learning~\cite{he2016deep, ren2015faster, he2017mask, goodfellow2014generative, croitoru2023diffusion}. These achievements have predominantly been constrained to supervised learning, reliant on abundant labeled data. However, collecting and annotating data from various domains is labor-expensive and time-consuming. 
To address this problem, Unsupervised Domain Adaptation (UDA)~\cite{pan2009survey, yang2020transfer} emerges as a promising solution. It alleviates the data annotation expenses and transfers knowledge from a labeled source domain to an unlabeled target domain. 

Previous endeavors primarily focus on alleviating the domain divergence via metric learning~\cite{LongC0J15, LongZ0J17, ZhuZW19, Kang0YH19}, adversarial learning~\cite{GaninUAGLLML16, XuZNLWTZ20, WuIE20}, and self-training~\cite{liang2020we, jiang2022prototypical, yue2021prototypical}. Another line of solutions~\cite{abs-2109-06165, lai2023padclip, abs-2204-07683, abs-2108-05988,rangwani2022closer} enhance the network capacity by transiting from convolutional neural networks, e.g., ResNet~\cite{he2016deep} to the vision transformer (ViT)~\cite{dosovitskiy2020image}. However, recent methods~\cite{zhu2023patch, abs-2109-06165, WangGZ23, liang2020we} often neglect the negative impact caused by the significant domain gap. For instance, CDTrans~\cite{abs-2109-06165} is limited by the effectiveness of cross-attention: it heavily relies on the quality of pseudo labels, rendering it less effective as the domain gap widens~\cite{zhu2023patch}. The reason behind this is the predominant reliance of current UDA techniques on visual backbones, potentially indicating a deficiency in attaining the requisite semantic richness~\cite{singha2023ad}.

\begin{figure}[t]
\centering
\includegraphics[width=0.8\linewidth]{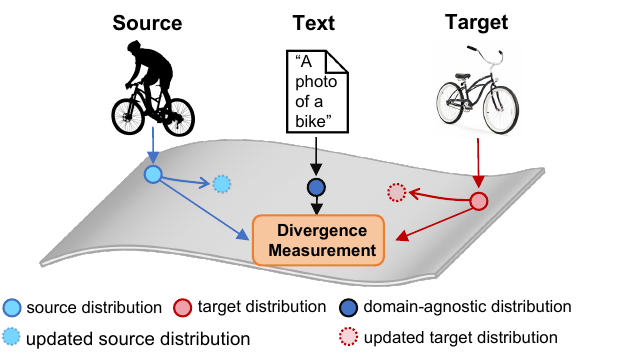}
\vspace{-10pt}
  \caption{\textbf{The key intuition of our CLIP-Div for measuring the domain divergence}. The domain-agnostic distribution, acquired through CLIP with language guidance, serves as a pivotal bridge between source and target domains through the design of divergence measurements.}
    \label{fig:intro}
    \vspace{-15pt}
\end{figure}

Large vision-language models (VLMs) have recently gained popularity due to their remarkable semantic richness and zero-shot generalization capability. For instance, CLIP~\cite{radford2021learning} is a scalable contrastive pre-training model for jointly learning text and image features. It leverages a vast corpus of 400 million image-text pairs~\cite{schuhmann2021laion}. The success of VLMs indicates that: models trained on a sufficiently extensive dataset possess the potential to mitigate the domain gap between source and target domains through the inherent domain knowledge~\cite{lai2023padclip}. This has inspired recent CLIP-based approaches~\cite{ge2022domain, singha2023ad, lai2023padclip, li2024split, du2024domain}, seeking to directly fine-tune CLIP's visual encoder or learn text prompts to the challenging UDA task. 
For instance, DAPL~\cite{ge2022domain} pioneered in utilizing CLIP for UDA tasks, especially focusing on disentangled semantic and domain representations through prompt learning. AD-CLIP~\cite{singha2023ad} tackles the UDA problem by leveraging the semantic richness of CLIP. PADCLIP~\cite{lai2023padclip} leverages the zero-shot generalization of CLIP for UDA and addresses the issue of catastrophic forgetting when fine-tuning CLIP on the target domain. 

In this paper, we shift the gear to a new direction by \textit{directly leveraging the inherent semantic richness and zero-shot generalization capability of CLIP to measure the domain divergence}, rather than fine-tuning and prompting to adapt CLIP to the target data. 
In light of this, we propose \textbf{CLIP-Div}, a novel language-guided UDA approach that harnesses CLIP for UDA. Our approach explores CLIP to serve two key purposes: \textbf{1}) measuring the \textit{domain divergence} via the domain-agnostic distribution derived from CLIP and \textbf{2}) calibrating the \textit{target pseudo labels with language guidance} to effectively reduce the domain divergence and improve the UDA model’s generalization capability.

Firstly, the key technical contribution of CLIP-Div is to propose two novel language-guided domain divergence measurement losses: \textbf{absolute divergence} and \textbf{relative divergence} (Sec.~\ref{lgdm}). Our intuition is that: \textit{compared with directly aligning source and target domains, reducing the gap between the two distributions and the domain-agnostic distribution -- as a crucial bridge -- can facilitate the domain alignment}, as shown in Fig.~\ref{fig:intro}.  These loss terms furnish precise guidelines for aligning the distributions of the source and target domains with the domain-agnostic distribution. Specifically, the absolute divergence pushes the distributions of two domains closer to the same domain-agnostic distribution with language guidance. To further enhance the domain alignment, the relative divergence ensures that the distance between a source sample and a target sample in the domain-specific distributions aligns with the distance between their corresponding samples in the domain-agnostic distribution. 
Secondly, we address the potential impact of less reliable target pseudo-labels in the presence of a substantial domain gap by proposing a \textbf{language-guided pseudo-labeling} strategy (Sec.~\ref{lgpl}). This approach aims to calibrate the target pseudo labels and subsequently employs self-training to bolster the UDA model's generalization capability on the target domain.


We conduct extensive experiments on four benchmark datasets, namely Office-31 \cite{SaenkoKFD10}, Office-Home \cite{VenkateswaraECP17}, VisDA-2017 \cite{abs-1710-06924}, and DomainNet \cite{PengBXHSW19}. The experimental results present that CLIP-Div significantly outperforms the state-of-the-art (SoTA) CNN-based approaches \cite{NaJCH21, lai2023padclip} by substantial margins:\textbf{ +10.3}\% on Office-Home, \textbf{+1.5}\% on Office-31, \textbf{+0.2}\% on VisDA-2017, and \textbf{+24.3}\% on DomainNet. 

In summary, our main contributions are: (\textbf{I}) We propose CLIP-Div, a novel language-guided approach for UDA, leveraging the domain-agnostic distribution derived from CLIP as a crucial bridge to facilitate the alignment between source and target domains; (\textbf{II}) We introduce two novel language-guided divergence measurement losses, namely the absolute divergence and relative divergence, furnishing guidance for aligning the two domains with the acquired domain-agnostic distribution; (\textbf{III}) We incorporate language-guided pseudo-labeling strategy tailored for target pseudo-label calibration, resulting in improved UDA model's generalization ability on the target domain through self-training; (\textbf{IV}) Extensive experiments validate the effectiveness of our proposed CLIP-Div, demonstrating SoTA results compared with prior methods.

\section{Related Work}
\label{sec:related work}

\textbf{Unsupervised Domain Adaptation (UDA).} 
Existing UDA works~\cite{zhu2024energy, zhu2023unified, pan2009survey, yang2020transfer} focus on aligning source and target domains or augmenting network capacity by transitioning from the convolutional neural networks, such as ResNet~\cite{he2016deep}, to ViT architectures~\cite{dosovitskiy2020image}. Recent advancements in domain alignment predominantly fall into three categories: 1) metric learning, 2) adversarial learning, and 3) self-training. Specifically, the first line of methods~\cite{LongC0J15, LongZ0J17, ZhuZW19, Kang0YH19} operates by quantifying domain divergence through various metrics, ultimately yielding domain-variant features. The second line of approaches~\cite{GaninUAGLLML16, XuZNLWTZ20, WuIE20}, on the other hand, probes to encourage samples from different domains to be non-discriminative via the adversarial loss, thereby acquiring domain-invariant representations.
A third line of approaches~\cite{liang2020we, liang2021domain, jiang2022prototypical, yue2021prototypical, DBLP:journals/corr/abs-2401-05014} involves pseudo-label assignment in the target domain, followed by network retraining in a supervised manner. A distinct line of innovative solutions, such as CDTrans~\cite{abs-2109-06165}, PMTrans~\cite{zhu2023patch}, and other ViT-based models~\cite{lai2023padclip, abs-2109-06165, abs-2204-07683, abs-2108-05988, rangwani2022closer}, capitalizes on ViT models to facilitate knowledge transfer. For example, 
CDTrans~\cite{abs-2109-06165} has a specific limitation where the effectiveness of cross-attention heavily relies on the quality of pseudo labels, rendering it less effective as the domain gap widens~\cite{zhu2023patch}.
\textit{Our CLIP-Div is a type of self-training approach; however, it directly leverages the remarkable semantic richness and zero-shot generalization capability of CLIP to acquire the domain-agnostic distribution as a pivotal bridge and then proposes two language-guided divergence measurement loss terms for aligning domains. }

\noindent\textbf{CLIP for UDA.} CLIP have demonstrated promising results in learning generic visual representations and facilitating zero-shot transfer to diverse downstream classification tasks through the use of
prompts~\cite{zhou2022conditional}. In particular, CLIP exhibits superior performance in solving UDA problems~\cite{ge2022domain, singha2023ad, lai2023padclip, li2024split, du2024domain}. DAPL~\cite{ge2022domain} pioneers the utilization of CLIP for the UDA task. It introduces a novel approach by designing domain-specific contexts for each domain, thereby promoting the learning of distinct domain representations. AD-CLIP~\cite{singha2023ad} tackles the UDA problem by leveraging the semantic richness of CLIP and learning domain-invariant and class-generic prompt tokens with visual space features. More recently, PADCLIP~\cite{lai2023padclip} addresses the challenge of catastrophic forgetting in fine-tuning large-scale pre-trained models while mitigating domain gaps. \textit{Differently, we aim to directly leverage the inherent zero-shot capability of CLIP, rather than fine-tuning and prompting, to construct the domain-agnostic distribution which can bridge the domain gaps, and calibrate the target pseudo labels for improving the UDA model's generalization ability on the target domain (as shown in Tab.~\ref{tab:backbone2}). } 

\noindent\textbf{Pseudo-labeling (PL)} involves assigning the highest anticipated probability to unlabeled data and subsequently utilizing the labeled data for effective fine-tuning~\cite{lee2013pseudo}. Initially introduced in  semi-supervised learning~\cite{cascante2021curriculum, zhang2021flexmatch, xiong2021multiview}, PL has gained prominence in UDA~\cite{mei2020instance, li2022cross,liu2021cycle}
. Specifically, PL entails labeling unlabeled target data with progressively refined high-confidence predictions, utilizing these pseudo labels as supervision in subsequent training loops. Some methods~\cite{zhang2018collaborative, choi2019pseudo, kim2022semi, sohn2020fixmatch, zhang2021flexmatch} directly employ pseudo labels as consistency regularization to enhance prediction consistency for unlabeled target data. Another line of works~\cite{long2017deep, long2018conditional, zou2018unsupervised} integrates target pseudo labels into the adaptation module to facilitate discriminative distribution alignment. However, these existing methods share the same assumption that the source and target domain distributions are similar, which may not hold in real-world scenarios. Such distribution shifts can lead to performance degradation in the self-training process. Therefore, \textit{we leverage the inherent domain knowledge of CLIP to calibrate target pseudo labels with language guidance, enhancing the UDA model's generalization ability on the target domain.}

\begin{figure*}[t]
\includegraphics[width=.98\linewidth]{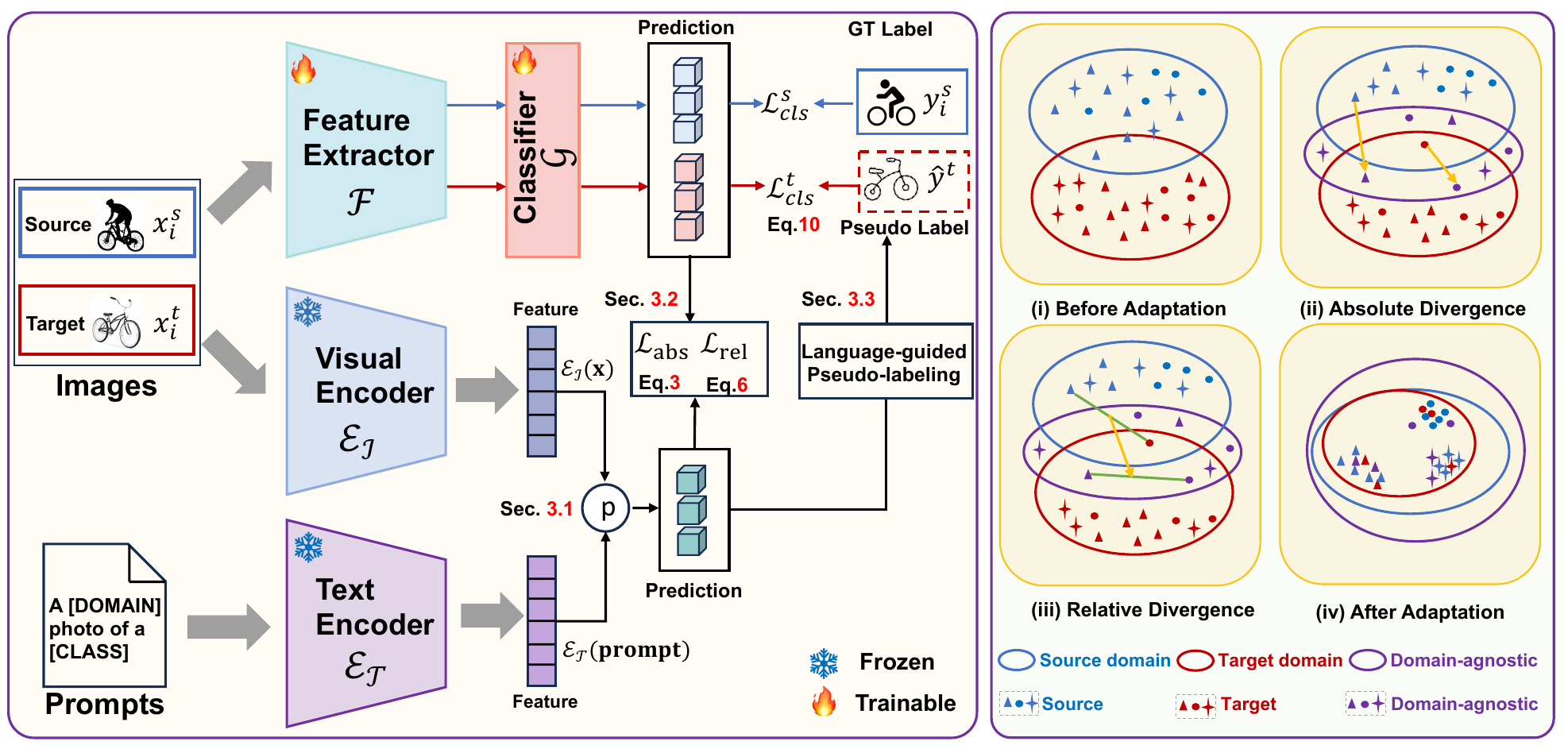}
\vspace{-10pt}
   \caption{(\textbf{Left}) \textbf{Overview of the pipeline of our proposed CLIP-Div}. The learned UDA model comprises a feature extractor $\mathcal{F}$ and a classifier $\mathcal{G}$. The encoders of frozen CLIP, denoted as $\mathcal{E_I}$ and $\mathcal{E_T}$, are harnessed for the acquisition of a domain-agnostic distribution. (\textbf{Right}) The intuition of our two proposed language-guided divergence measurement losses and their collaborative effects. (i) In UDA, the objective is to align source and target domains by learning domain-invariant representations; (ii) the absolute divergence measurement loss works to bring the distributions of both domains closer to the domain-agnostic distribution; (iii) the relative divergence measurement loss facilitates the domain alignment based on the distance between samples; (iv) collectively, these language-guided divergence measurement losses contribute to mitigating the domain gap between the source and target domains. }
   \label{fig:framework}
   \vspace{-10pt}
\end{figure*}

\section{Method}
\label{sec:method}
In this paper, we denote the source domain dataset as $\mathcal{D}^s=\left\{\left(x_i^s, y_i^s\right)\right\}_{i=1}^{n_s}$, comprising $n_s$ labeled samples, where $y \in \mathcal{Y} \subseteq \mathbb{R}^K$ represents the one-hot ground-truth label and $K$ signifies the total number of classes within the label set $\mathcal{C}=\{1,2, \cdots, K\}$. Concurrently, the target domain dataset, denoted as $\mathcal{D}^t=\left\{\left(x_i^t\right)\right\}_{i=1}^{n_t}$, consists of $n_t$ unlabeled samples, sharing the same label set $\mathcal{C}$ as that of $\mathcal{D}^s$. In the UDA scenario, we have access to the UDA model denoted as $\mathcal{G}(\mathcal{F}(\cdot))$. Here, $\mathcal{F}$ represents the feature extractor, which is followed by a linear classifier $\mathcal{G}$. 
The primary objective is to harness the recent advances in large pre-trained vision-language models, such as CLIP, to mitigate the domain gap and enhance the UDA model's generalization capability on the target domain. To achieve this, our approach comprises several key components. Firstly, we introduce the utilization of CLIP for addressing the UDA task (Sec.~\ref{clip_intro}). Subsequently, we introduce two language-guided divergence measurement loss terms to reduce the domain divergence (Sec.~\ref{lgdm}). Finally, we devise a language-guided pseudo-labeling strategy to calibrate the target pseudo labels, thereby enhancing the UDA model's generalization performance on the target domain (Sec.~\ref{lgpl}). The overview of our proposed CLIP-Div is illustrated in the left portion of Fig.~\ref{fig:framework}.

\subsection{Preliminaries}
\label{clip_intro}
Our initial step involves adapting CLIP to suit the requirements of UDA tasks. CLIP comprises a visual encoder, denoted as $\mathcal{E_I}$, responsible for mapping images into low-dimensional image representations, and a text encoder, $\mathcal{E_T}$, which maps text sentences into text representations. In accordance with previous prompt engineering works~\cite{radford2021learning, lai2023padclip}, we prepare image-text pairs tailored for UDA datasets. To structure these pairs, we employ a designated format, where $\boldsymbol{prompt}$ represents a sentence of the form ``a [\textit{DOMAIN}] photo of a [\textit{CLASS}]’’. Here, [\textit{CLASS}] denotes the name of a classification category, and [\textit{DOMAIN}] signifies the domain name in the specific UDA task, such as ``an art photo of a bike’’. We denote prompts associated with the source and target data as $\boldsymbol{prompt}^{s}$ and $\boldsymbol{prompt}^{t}$, respectively. Furthermore, we introduce $\boldsymbol{prompt}^{a}$ to represent a domain-agnostic prompt structured as ``a photo of a [\textit{CLASS}]’’. Note that $\boldsymbol{prompt}^s_k$, $\boldsymbol{prompt}^t_k$, and $\boldsymbol{prompt}^a_k$ are tailored for images associated with the $k$-th category, respectively. To derive the data distributions within the logit space, we adhere to the CLIP zero-shot inference methodology. This involves comparing the image representations extracted by visual encoder $\mathcal{E_I}$ with classification weights generated by the text encoder $\mathcal{E_T}$. By utilizing $K$ descriptions corresponding to $K$ classes within specific domains, we calculate the probability that a given training image belongs to the $k$-th category.

{\setlength\abovedisplayskip{4pt}
\setlength\belowdisplayskip{2pt}
\begin{equation}
\small
   {p}_k(\boldsymbol{x}, \boldsymbol{prompt})=\frac{\exp \left(\cos \left(\mathcal{E_{T}}({\boldsymbol{prompt}}_{k}), \mathcal{E_{I}}\left(\boldsymbol{x}\right) / \tau \right)\right.}{\sum_{k=1}^K \exp \left(\cos \left(\mathcal{E_{T}}({\boldsymbol{prompt}}_{k}), \mathcal{E_{I}}\left(\boldsymbol{x}\right) / \tau \right)\right.}, 
   \label{eq:clip}
\end{equation}}

where $\tau$ is the temperature parameter, and $cos$ refers to cosine similarity~\cite{radford2021learning}. We represent a vector of probabilities $p_k$ as $p$ within the logit space.

\subsection{Language-guided Divergence Measurement}
\label{lgdm}
Traditional UDA approaches predominantly rely on metric-based learning~\cite{LongC0J15, LongZ0J17, ZhuZW19, Kang0YH19}, employing only visual representations to measure the domain divergence. However, these methods frequently lack semantic richness, leading to sub-optimal performance in critical scenarios~\cite{singha2023ad}. To address this limitation, we harness CLIP to establish an optimal domain-agnostic distribution, incorporating distinct textural prompts. Subsequently, the domain-agnostic distribution serves as a pivotal bridge between source and target domains. Specifically, our objective is to align source and target domains by guiding their domain-specific distributions toward the domain-agnostic distribution. \textit{Once the alignment between each domain-specific distribution and the domain-agnostic distribution is achieved, the two domains are accordingly aligned}. Building upon this rationale, we introduce two novel language-guided divergence measurement loss terms: absolute divergence and relative divergence. These loss terms are described in detail below.

\noindent\textbf{Absolute Divergence.}
 In the right portion of Fig.~\ref{fig:framework}, we introduce two absolute divergence losses for the source and target domains. Initially, we leverage CLIP to acquire the domain-agnostic distribution in the logit space with domain-agnostic prompts $\boldsymbol{prompt}^{a}$. This entails inputting source data $\boldsymbol{x}^{s}$ and target data $\boldsymbol{x}^{t}$ into the visual encoder $\mathcal{E_I}$, generating visual feature representations $\mathcal{E_I(}\boldsymbol{x}^{s})$ and $\mathcal{E_I}(\boldsymbol{x}^{t})$. Simultaneously, domain-agnostic prompts $\boldsymbol{prompt}^a$ (e.g., a photo of a bike) are input into the text encoder $\mathcal{E_T}$ to obtain text feature representations $\mathcal{E_T}(\boldsymbol{prompt}^{a})$. The probabilities ${p}(\boldsymbol{x}, \boldsymbol{prompt}^{a})$ for source and target data are then obtained using Eq.\ref{eq:clip}, yielding the domain-agnostic distribution in the logit space. For the UDA model $\mathcal{G}(\mathcal{F}(\cdot))$, we input the same source data $\boldsymbol{x}^{s}$ and target data $\boldsymbol{x}^{t}$ to obtain probabilities $\mathcal{G}(\mathcal{F}(x))$, which yields domain-specific distributions in the logit space. 

To encourage the UDA model $\mathcal{G}(\mathcal{F}(\cdot))$ to learn the domain-invariant distribution, we strive to align the domain-specific distributions with the domain-agnostic distribution. To quantify the divergence between domain-specific distributions and the domain-agnostic distribution, an absolute divergence is introduced as
\begin{equation}
\begin{aligned}
    \mathcal{L}_{abs}^{s} &= \mathbb{E}_{\boldsymbol{x}^{s} \sim D^{s}} KL({p}(\boldsymbol{x}^{s}, \boldsymbol{prompt}^{a}) \|\mathcal{G}(\mathcal{F}(\boldsymbol{x}^{s}))),\\
    \mathcal{L}_{abs}^{t} &= \mathbb{E}_{\boldsymbol{x}^{t} \sim D^{t}} KL({p}(\boldsymbol{x}^{t}, \boldsymbol{prompt}^{a}) \| \mathcal{G}(\mathcal{F}(\boldsymbol{x}^{t}))),
\end{aligned}
\end{equation}
where $KL(\cdot)$ represents the Kullback-Leibler divergence. Consolidating the aforementioned divergence measurement losses for both source and target domains, we formulate the function for total absolute divergence as
\begin{equation}
    \mathcal{L}_{abs} = \mathcal{L}_{abs}^{s}+\mathcal{L}_{abs}^{t}.
\end{equation}
Minimizing the absolute divergence $\mathcal{L}_{abs}$ serves to narrow the gap between domain-specific distributions and the domain-agnostic distribution, thereby promoting effective alignment between the source and target domains.

\noindent\textbf{Relative Divergence.} 
While the absolute divergence effectively aids in mitigating domain gaps, optimizing the UDA model $\mathcal{G}(\mathcal{F}(\cdot))$ to identify optimal solutions is not straightforward. To address this challenge, we further introduce a novel language-guided relative divergence loss to facilitate the domain alignment (Fig.~\ref{fig:framework} right). The underlying principle is that if the distance between a source sample and a target sample in the domain-specific distributions aligns with the distance between their corresponding samples in the domain-agnostic distribution, the domain-specific distributions are in closer proximity to the domain-agnostic distribution. This alignment eventually leads to a decrease in the divergence between two domains.

To achieve this and \textit{fully explore the inherent knowledge of CLIP}, we craft domain-specific prompts tailored for both source and target domains, facilitating the construction of the domain-agnostic distribution (e.g., a clipart photo of a bike). Initially, we adopt the averaged representation of domain-specific text prompts as the domain-agnostic text representations. The domain-agnostic text representation for class $k$ is

{\setlength\abovedisplayskip{6pt}
\setlength\belowdisplayskip{2pt}
\begin{equation}
\mathcal{E_{T}}({\boldsymbol{prompt}}^{avg}_{k}) = \frac{\mathcal{E_{T}}({\boldsymbol{prompt}}_{k}^s)+\mathcal{E_{T}}({\boldsymbol{prompt}}_{k}^t)}{2}.
\end{equation}}

Subsequently, relying on the domain-agnostic text representations, we formulate the probabilities for source and target samples as
{\setlength\abovedisplayskip{4pt}
\setlength\belowdisplayskip{2pt}
\begin{equation}
\small
{p}_k(\boldsymbol{x}, \boldsymbol{prompt}^{avg})=\frac{\exp \left(\cos \left(\mathcal{E_{T}}({\boldsymbol{prompt}}^{avg}_{k}), \mathcal{E_{I}}\left(\boldsymbol{x}\right) / \tau \right)\right.}{\sum_{k=1}^K \exp \left(\cos \left(\mathcal{E_{T}}({\boldsymbol{prompt}}^{avg}_{k}), \mathcal{E_{I}}\left(\boldsymbol{x}\right) / \tau \right)\right.},  
\end{equation}}

To equate the distances within the domain-specific distributions to those in the domain-agnostic distribution, we define the relative divergence loss as
\begin{equation}
\begin{aligned}
    \Delta_{1}&={p}(\boldsymbol{x}^{s}, \boldsymbol{prompt}^{avg})-{p}(\boldsymbol{x}^{t}, \boldsymbol{prompt}^{avg}), \\
    \Delta_{2}&=\mathcal{G}(\mathcal{F}(\boldsymbol{x}^{s}))-\mathcal{G}(\mathcal{F}(\boldsymbol{x}^{t})),\\
    \mathcal{L}_{rel}&=1-\frac{\Delta_1\cdot \Delta_{2}}{|\Delta_1||\Delta_2|},
\end{aligned}
\end{equation}
where $\Delta_1$ denotes the distance between a source sample and a target sample within the domain-agnostic distribution. Correspondingly, $\Delta_2$ is derived as the distance between their corresponding samples within the domain-specific distribution. Minimizing the distance indicates the enhanced proximity of the domain-specific distributions to the domain-agnostic distribution, ultimately resulting in the complete alignment of the two domains.

\subsection{Language-guided Pseudo-Labeling}
\label{lgpl}
Prior self-training based UDA works~\cite{mei2020instance, li2022cross,liu2021cycle} predominantly rely on pseudo labels from target data for supervision in subsequent training loops, showcasing commendable performance. However, the substantial domain gap between the source and target domains introduces challenges, rendering the pseudo labels of target data less reliable. This unreliability, in turn, results in sub-optimal classification performance and adverse effects on the UDA model's generalization capability with self-training. Motivated by the zero-shot generalization and transfer capability inherent in CLIP, we propose a novel strategy: language-guided pseudo-labeling. This strategy aims to refine target pseudo labels, subsequently leveraging these refined target labels to retrain the model in a supervised manner. 

Building upon the pseudo-labeling methodology introduced in~\cite{liang2020we}, we now elucidate the language-guided pseudo-labeling strategy. 
First, we compute the centroid $c_k^t$ for each class in the target domain using the target features $\mathcal{F}\left(\boldsymbol{x}^{t}\right)$.
\begin{equation}
  c_k^t=\frac{\sum_{\boldsymbol{x}^{t} \sim D^{t}} (\delta_k^t+{p}_k^t(\boldsymbol{x}^t, \boldsymbol{prompt}^t)) \mathcal{F}\left(\boldsymbol{x}^{t}\right)}{\sum_{\boldsymbol{x}^{t} \sim D^{t}} (\delta_k^t+{p}_k^t(\boldsymbol{x}^t,\boldsymbol{prompt}^t))},
\end{equation}
where $\delta_k^t$ is the probability $\mathcal{G}(\mathcal{F}(\boldsymbol{x}^{t}))$ of target data $\boldsymbol{x}^{t}$ on class $k$. ${p}_k^t(\boldsymbol{x}^t,\boldsymbol{prompt}^t)$ represents the $k$-th category probability of target data $\boldsymbol{x}^{t}$ from CLIP with target domain-specific prompts $\boldsymbol{prompt}^t$. Leveraging CLIP's exceptional zero-shot generalization capability, we utilize the output probabilities obtained with the target domain-specific prompts to calibrate the centroids and enhance the reliability of centroids for target data. Subsequently, the pseudo labels for target data $\boldsymbol{x}^{t}$ are derived through the nearest centroid classifier:



\begin{equation}
  \hat{y}^t=\arg \min _k d\left(\mathcal{F}\left(\boldsymbol{x}^{t}\right), c_k^t\right),
\end{equation}

\begin{table*}[t]
\caption{Comparison with SoTA methods on Office-Home.}
\centering
\resizebox{0.9\linewidth}{!}{
\begin{tabular}{|c||l|lllllllllllll|}
\hline
Method &Publication&A$\to$ C& A$\to$ P & A $\to$ R & C $\to$ A &C $\to$ P &C $\to$ R &P$\to$ A& P$\to$ C & P$ \to$ R & R$ \to$ A &R$ \to$ C &R$ \to$ P& Avg     \\
\hline
ResNet-50~\cite{he2016deep} &CVPR'16& 44.9& 66.3 &74.3 &51.8& 61.9 &63.6& 52.4 &39.1& 71.2& 63.8& 45.9& 77.2 &\colorbox{lightgray}{59.4}   \\
MCD~\cite{SaitoWUH18} &CVPR'18&48.9& 68.3& 74.6& 61.3& 67.6& 68.8& 57.0& 47.1 &75.1& 69.1& 52.2& 79.6& \colorbox{lightgray}{64.1}\\
MDD~\cite{li2020maximum} &TPAMI' 20&54.9& 73.7& 77.8& 60.0& 71.4& 71.8& 61.2& 53.6& 78.1& 72.5& 60.2& 82.3& \colorbox{lightgray}{68.1}\\
BNM~\cite{CuiWZLH020}&CVPR'20&56.7 &77.5& 81.0& 67.3& 76.3& 77.1& 65.3& 55.1& 82.0 &73.6 &57.0& 84.3& \colorbox{lightgray}{71.1}\\
DALN~\cite{chen2022reusing} &CVPR'22& 57.8& 79.9& 82.0 &66.3& 76.2& 77.2& 66.7& 55.5& 81.3& 73.5& 60.4& 85.3& \colorbox{lightgray}{71.8}  \\
SDAT~\cite{rangwani2022closer} &PMLR'22& 58.2& 77.1& 82.2 &66.3& 77.6& 76.8& 63.3& 57.0& 82.2& 74.9& 64.7& 86.0& \colorbox{lightgray}{72.2}  \\
FixBi~\cite{NaJCH21} &CVPR'21& 58.1& 77.3& 80.4 &67.7& 79.5& 78.1& 65.8& 57.9& 81.7& 76.4& 62.9& 86.7& \colorbox{lightgray}{72.7}  \\
kSHOT~\cite{sun2022prior} &CVPR'22& 58.2& 80.0& 82.9 &71.1& 80.3& 80.7& 71.3& 56.8& 83.2& 75.5& 60.3& 86.6& \colorbox{lightgray}{73.9}  \\
\hline
DAPL~\cite{ge2022domain} &Arxiv'22&54.1 &84.3 &84.8& 74.4 &83.7 &85.0& 74.5& 54.6& 84.8& 75.2& 54.7 &83.8& \colorbox{lightgray}{74.5}\\
AD-CLIP~\cite{singha2023ad}&ICCV'23 & 55.4& 85.2& 85.6 &76.1 &85.8& 86.2 &76.7& 56.1& 85.4 &76.8& 56.1& 85.5& \colorbox{lightgray}{75.9} \\
PADCLIP~\cite{lai2023padclip}&ICCV'23 &57.5& 84.0& 83.8& 77.8 &85.5& 84.7 &76.3 &59.2& 85.4 &78.1& 60.2& 86.7& \colorbox{lightgray}{76.6}\\
UniMoS~\cite{li2024split}&CVPR'24&59.5 &89.4& 86.9 &75.2 &89.6& 86.8 &75.4& 58.4 &87.2& 76.9& 59.5& 89.7 &\colorbox{lightgray}{77.9}\\
DAMP~\cite{du2024domain}&CVPR'24& 59.7& 88.5 &86.8 &76.6& 88.9& 87.0 &76.3 &59.6& 87.1& 77.0& 61.0 &89.9& \colorbox{lightgray}{78.2}\\
\hline
\textbf{CLIP-Div (ResNet-50)}&-&57.2&80.4&82.9&73.9&80.7&81.1&72.8&58.6&83.5&73.3&59.9&81.9&\colorbox{lightgray}{73.9}\\
\textbf{CLIP-Div (ResNet-101)}&-&61.3&84.1&85.1&77.1&82.8&85.0&75.6&63.1&85.5&75.9&63.7&84.5&\colorbox{lightgray}{77.0}\\
\textbf{CLIP-Div (ViT-B)}&-&71.4&89.5&89.4&83.0&89.9&89.0&81.5&73.0&89.4&81.8&73.9&90.2&\colorbox{lightgray}{83.5}\\
\textbf{CLIP-Div (ViT-L)} &-& \textbf{79.1}&\textbf{93.7}& \textbf{93.3}& \textbf{87.7}& \textbf{94.0}&\textbf{ 93.2}& \textbf{86.8 }&\textbf{80.6} &\textbf{93.4} &\textbf{86.8}& \textbf{80.1}& \textbf{93.4}& \colorbox{lightgray}{\textbf{88.5}}\\
\hline
\end{tabular}}
\label{tab:officehome}
\vspace{-15pt}
\end{table*}

Finally, we calculate the target centroids utilizing the newly assigned pseudo labels as

\begin{equation}
   \begin{aligned}
{c_k^t}^{\prime} & =\frac{\sum_{\boldsymbol{x}^{t} \sim D^{t}} \mathbb{I}\left(\hat{y}^t=k\right) \mathcal{F}\left(\boldsymbol{x}^t\right)}{\sum_{\boldsymbol{x}^{t} \sim D^{t}} \mathbb{I}\left(\hat{y}^t=k\right)}, \\
{\hat{y}^t} & =\arg \min _k d\left(\mathcal{F}\left(\boldsymbol{x}^{t}\right), {c_k^t}^{\prime} \right),
\end{aligned} 
\label{Eq}
\end{equation}
where $d(\cdot)$ represents the cosine distance. The term $\hat{y}^t$ is coined as target pseudo labels, derived from centroids obtained in an unsupervised manner. Eq.~\ref{Eq} can be iteratively updated for multiple rounds, and this work only employs a single round. Building upon the pseudo labels $\hat{y}^t$, the classification loss for the target data is defined as

\begin{equation}
\mathcal{L}_{cls}^{t}=\mathbb{E}_{\left(\boldsymbol{x}^{t}, \hat{\boldsymbol{y}}^{t}\right) \sim D^{t}} \ell\left(\mathcal{G}\left(\mathcal{F}\left(\boldsymbol{x}^{t}\right)\right), \hat{\boldsymbol{y}}^{t}\right),
\end{equation}
where $\ell$ is the cross entropy loss.
\subsection{Total Objective}

Connecting all the pieces above, the total objective of CLIP-Div is formulated as 
\begin{equation}
\mathcal{L}= \mathcal{L}_{cls}^{s}+\lambda_{abs}*\mathcal{L}_{abs}+\lambda_{rel}*\mathcal{L}_{rel}+\lambda_{pl}*\mathcal{L}_{pl},
\end{equation}

where $\lambda_{abs}$, $\lambda_{rel}$, and $\lambda_{pl}$ denote three hyper-parameters responsible for balancing each loss term. Additionally, $\mathcal{L}_{cls}^{s}$ represents the classification loss for the labeled source data. The specific values assigned to the hyper-parameters are $\lambda_{abs}=10$, $\lambda_{rel}=1$, and $\lambda_{pl}=0.1$, respectively.

\section{Experiment}
\subsection{Datasets and Implementation}
\label{sec:experiment}


We evaluate the effectiveness of our proposed CLIP-Div through extensive experiments on four prominent UDA benchmarks, including Office-Home \cite{VenkateswaraECP17}, Office-31 \cite{SaenkoKFD10}, VisDA-2017 \cite{abs-1710-06924}, and DomainNet \cite{PengBXHSW19}. \textbf{Office-Home} dataset encompasses four domains, each of which includes images from 65 categories, yielding a total of approximately 15,500 images. These domains consist of Art (\textbf{A}), Clipart (\textbf{C}), Product (\textbf{P}) and Real-World (\textbf{R}). \textbf{Office-31} comprises three domains: Amazon (\textbf{A}), DSLR (\textbf{D}), and Webcam (\textbf{W}), encompassing 31 categories. \textbf{VisDA-2017} is a large-scale benchmark including 152,397 synthetic images and 55,388 real images. We focus on the challenging synthetic-to-real transfer task. \textbf{DomainNet} is by far the largest dataset for domain adaptation, including approximately 0.6 million images distributed of 345 categories from six distinct domains: Clipart (\textbf{clp}), Infograph (\textbf{inf}), Painting (\textbf{pnt}), Quickdraw (\textbf{qdr}), Real-world (\textbf{rel}), and Sketch (\textbf{skt}). We build 12, 6, 1, and 30 transfer tasks on Office-Home, Office-31, VisDA-2017, and DomainNet, respectively.

In all experiments, we utilize ResNet as a feature extractor with a batch size of 32. The pre-trained ResNet-50 models serve as the backbone for all datasets, except ResNet-101 for VisDA-2017. The learning rates are set to 2e$-3$ for the feature extractor and 2e$-2$ for the classifier. Training is conducted on Office-Home and Office-31 datasets for 200 epochs, and on VisDA-2017 and DomainNet for 50 epochs. Specifically, mini-batch stochastic gradient descent with a momentum of 0.9 and the learning rate annealing strategy from Revgrad~\cite{ganin2015unsupervised} are employed. The learning rate is dynamically adjusted during SGD using the formula $\eta_\theta=\frac{\eta_0}{(1+\alpha \theta)^\beta}$, where $\theta$ linearly progresses from 0 to 1, $\eta_0=0.01$, $\alpha=10$, and $\beta=0.75$. Notably, during training, the visual encoder and text encoder within CLIP are frozen. CLIP-Div (ViT-B) employs ViT-B (patch size 16$\times$16) as the vision backbone, while CLIP-Div (ViT-L) utilizes ViT-L (patch size 14$\times$14). The training duration spans 100, 200, 50, and 50 epochs for Office-Home, Office-31, VisDA-2017, and DomainNet, respectively. Notably, we freeze the visual encoder and text encoder within CLIP during training, and \textbf{only utilize the ResNet backbone exclusively during inference}.

\begin{table*}[!t]
\caption{Comparison with SoTA methods on Office-31.}
\centering
\begin{tabular}{|c||l|lllllll|}
\hline
Method  &Publication&A $\to$ W & D$\to$ W & W$\to$ D & A$\to$ D &D$\to$ A &W$\to$ A & Avg \\
\hline
ResNet-50~\cite{he2016deep} &CVPR'16  &  68.9& 68.4 &62.5& 96.7& 60.7& 99.3& \colorbox{lightgray}{76.1}   \\
BNM~\cite{CuiWZLH020} &CVPR'20&91.5&98.5&\textbf{100.}&90.3&70.9&71.6&\colorbox{lightgray}{87.1}\\
MDD~\cite{li2020maximum}&TPAMI'20&94.5& 98.4& \textbf{100.}& 93.5& 74.6 &72.2& \colorbox{lightgray}{88.9}\\
SCDA~\cite{0008XLLLQL21}&ICCV'21&94.2 &98.7& 99.8& 95.2& 75.7& 76.2& \colorbox{lightgray}{90.0}\\
DALN~\cite{chen2022reusing}&CVPR'22&  95.2& 99.1& \textbf{100.}& 95.4&76.4& 76.5& \colorbox{lightgray}{90.4}  \\
kSHOT~\cite{sun2022prior}&CVPR'22&  \textbf{98.5}& 99.0& 99.8& \textbf{97.6}&75.0& 76.2& \colorbox{lightgray}{91.0}  \\
FixBi~\cite{NaJCH21}&CVPR'21&  96.1& \textbf{99.3}& \textbf{100.}& 95.0&78.7& 79.4& \colorbox{lightgray}{91.4}  \\
SDAT~\cite{rangwani2022closer}&PMLR'22&  97.2& 99.0& 99.8& 95.0&78.0& 79.4& \colorbox{lightgray}{91.4}  \\
\hline
\textbf{CLIP-Div (ResNet-50)}&-&87.6&97.5&98.8&87.8&74.8&76.3&\colorbox{lightgray}{87.1}\\
\textbf{CLIP-Div (ResNet-101)}&-&88.6&97.9&99.2&90.4&75.3&76.8&\colorbox{lightgray}{88.0}\\
\textbf{CLIP-Div (ViT-B)}&-&91.8&98.6&99.6&92.8&80.4&81.5&\colorbox{lightgray}{90.8}\\
\textbf{CLIP-Div (ViT-L)}&-&93.7&98.4&99.5&94.8&\textbf{84.7}&\textbf{85.4}&\colorbox{lightgray}{\textbf{92.9}}\\
\hline
\end{tabular}
\label{tab:office31_res}
\vspace{-10pt}
\end{table*}

\subsection{Comparisons with Prior Works}
We conduct a comparative analysis, pitting our CLIP-Div against SoTA CNN-based approaches in the realm of UDA. The compared methods include FixBi \cite{NaJCH21}, MCD \cite{SaitoWUH18}, CGDM~\cite{du2021cross}, MDD~\cite{li2020maximum},  SWD \cite{LeeBBU19}, SCDA \cite{0008XLLLQL21}, BNM~\cite{CuiWZLH020}, DALN~\cite{chen2022reusing}, SPA \cite{xiao2023spa}, SDAT \cite{rangwani2022closer}, kSHOT~\cite{sun2022prior} as well as DAPL \cite{ge2022domain}, AD-CLIP~\cite{singha2023ad}, PADCLIP~\cite{lai2023padclip}, UniMoS~\cite{li2024split}, and DAMP~\cite{du2024domain}. Note that DAPL~\cite{ge2022domain}, AD-CLIP~\cite{singha2023ad}, PADCLIP~\cite{lai2023padclip}, UniMoS~\cite{li2024split}, and DAMP~\cite{du2024domain} harness CLIP with ResNet backbone to tackle the complex challenges presented by UDA. CLIP-Div (ResNet-50, ResNet-101, ViT-B, or ViT-L) utilizes CLIP with vision backbone (ResNet-50, ResNet-101, ViT-B, or ViT-L) to learn the domain-agnostic distribution. Note that the best performance is marked as \textbf{bold} in the result tables.

\noindent \textbf{Results on Office-Home.} In Tab.\ref{tab:officehome}, we compare CLIP-Div with recent UDA methods on Office-Home. Notably, CLIP-Div (ViT-L) demonstrates robust performance \textbf{across all tasks}, outperforming both CNN-based and CLIP-based methods. Particularly noteworthy are its substantial improvements, exhibiting approximately $\textbf{19.4}\%$ and $\textbf{21.0}\%$ higher accuracy compared to SoTA methods in the transfer tasks A $\to$ C and P $\to$ C, respectively. With an average accuracy of $\textbf{88.5}\%$, CLIP-Div (ViT-L) showcases a $\textbf{10.3}\%$ enhancement over DAMP~\cite{du2024domain}. Note that utilizing the domain-agnostic distribution derived from CLIP (ResNet-50 or ResNet-101) to bridge domains results in slightly inferior performance compared to CLIP-based methods. This performance discrepancy arises because an unreliable domain-agnostic distribution leads to suboptimal domain alignment. Furthermore, CLIP-based methods, such as AD-CLIP ~\cite{singha2023ad} and PADCLIP~\cite{lai2023padclip}, employ prompt learning or fine-tuning of the CLIP visual encoder to reduce domain divergence, which can be computationally intensive. Detailed analysis and further discussion will be provided below.


\noindent \textbf{Results on Office-31.} Tab.~\ref{tab:office31_res} presents a comprehensive comparison of CLIP-Div against recent UDA methods on the Office-31 dataset. The results show CLIP-Div (ViT-L) outperforms SoTA methods with \textbf{1.5}\% average accuracy boost and achieves \textbf{92.9}\% average accuracy. Note that CLIP-Div (ViT-L) achieves an \textbf{6.7}\% accuracy boost in the D $\to$ A transfer task and an \textbf{6.0}\% improvement in W $\to$ A. However, CLIP-Div (ViT-L) exhibits sub-optimal performance in tasks A $\to$ D and A $\to$ W. This can be attributed to CLIP-Div aligning distributions through \textit{the utilization of the sub-optimal domain-agnostic distribution} derived from CLIP, as 
evident in Tab.~\ref{tab:office31_clip}. Upon closer inspection of Tab.~\ref{tab:office31_clip}, CLIP-Div ( ViT-B) outperforms CLIP (ViT-B) in target tasks A, D, and W with \textbf{4.1}\%,\textbf{ 17.1}\%, and \textbf{18.2}\% accuracy augmentation, respectively. The comparable performance over prior methods, coupled with the superiority over CLIP, systematically validates the effectiveness of CLIP-Div and \textit{substantiates the soundness of our intuition} in Fig.~\ref{fig:intro}. 

\begin{table}[t]
\caption{Comparison with CLIP on Office-31.}
\centering
\resizebox{0.99\linewidth}{!}{ 
\begin{tabular}{|c||llll|}
\hline
Method&A&D&W&Avg \\
\hline
CLIP (ResNet-50)&68.3& 66.7& 64.0&\colorbox{lightgray}{66.3}\\
CLIP-Div (ResNet-50)&75.6$_{\textcolor{red}{+7.3}}$&93.3$_{\textcolor{red}{+26.6}}$&92.6$_{\textcolor{red}{+28.6}}$&\colorbox{lightgray}{87.2}$_{\textcolor{red}{+20.9}}$\\
\hline
CLIP (ResNet-101)&70.3&73.3&73.1&\colorbox{lightgray}{72.2}\\
CLIP-Div (ResNet-101)&76.1$_{\textcolor{red}{+5.8}}$&94.8$_{\textcolor{red}{+21.5}}$&93.3$_{\textcolor{red}{+20.2}}$&\colorbox{lightgray}{88.1}$_{\textcolor{red}{+15.9}}$\\
\hline
CLIP (ViT-B)&76.9&79.1&77.0&\colorbox{lightgray}{77.7} \\
CLIP-Div (ViT-B)&$81.0_{\textcolor{red}{+4.1}}$&$96.2_{\textcolor{red}{+17.1}}$&$95.2_{\textcolor{red}{+18.2}}$&\colorbox{lightgray}{90.8}$_{\textcolor{red}{+13.1}}$\\
\hline
CLIP (ViT-L)&82.4&84.5&85.4&\colorbox{lightgray}{84.1}\\
CLIP-Div (ViT-L)&85.1$_{\textcolor{red}{+2.7}}$&97.2$_{\textcolor{red}{+12.7}}$&96.1$_{\textcolor{red}{+10.7}}$&\colorbox{lightgray}{92.8}$_{\textcolor{red}{+8.7}}$\\
\hline
\end{tabular}}
\label{tab:office31_clip}
\vspace{-10pt}
\end{table}


\noindent \textbf{Results on VisDA-2017.} Tab.~\ref{tab:VisDA1} presents the classification accuracy results on the VisDA-2017 dataset based on ResNet-101. Our CLIP-Div (ViT-L) achieves SoTA performance with an impressive average accuracy of \textbf{88.7}\%. CLIP-Div (ViT-L) excels with the highest accuracy in\textbf{ 4 }categories, underscoring the effectiveness of CLIP-Div with language guidance.

\begin{table*}[t]
\caption{Comparison with SoTA methods on VisDA-2017.}
\centering
\begin{tabular}{|c||l|lllllllllllll|}
\hline
Method &Publication&plane& bcycl& bus &car& horse& knife& mcycl& person& plant &sktbrd &train& truck& Avg     \\
\hline
ResNet-101~\cite{he2016deep}&CVPR'16 & 55.1 &53.3& 61.9& 59.1& 80.6 &17.9 &79.7& 31.2& 81.0 &26.5 &73.5& 8.5& \colorbox{lightgray}{52.4} \\
BNM~\cite{CuiWZLH020} &CVPR'20&89.6 &61.5 &76.9 &55.0& 89.3& 69.1 &81.3 &65.5& 90.0 &47.3& 89.1 &30.1& \colorbox{lightgray}{70.4}\\
MCD~\cite{SaitoWUH18}&CVPR'18&  87.0 &60.9 &83.7& 64.0& 88.9& 79.6& 84.7& 76.9& 88.6& 40.3& 83.0& 25.8 &\colorbox{lightgray}{71.9}\\
SWD~\cite{LeeBBU19}&CVPR'19&90.8 &82.5 &81.7 &70.5& 91.7 &69.5& 86.3& 77.5 &87.4& 63.6& 85.6 &29.2 &\colorbox{lightgray}{76.4}\\
SDAT~\cite{rangwani2022closer}&PMLR'22 & 95.8 &85.5& 76.9& 69.0 &93.5& 97.4 &88.5& 78.2& 93.1&91.6& 86.3 &55.3& \colorbox{lightgray}{84.3}  \\
kSHOT~\cite{sun2022prior} &CVPR'22& 95.7 &88.7& 81.4& 73.4 &94.7& 94.2 &88.1& 82.5& 93.4&91.1& 87.2 &63.1& \colorbox{lightgray}{86.1}  \\
FixBi~\cite{NaJCH21} &CVPR'21& 96.1 &87.8& 90.5& \textbf{90.3} &96.8& \textbf{95.3} &92.8& \textbf{88.7}& \textbf{97.2} &94.2& 90.9 &25.7& \colorbox{lightgray}{87.2}  \\
\hline
DAPL~\cite{ge2022domain}&Arxiv'21&97.8& 83.1 &88.8& 77.9 &\textbf{97.4}& 91.5 &94.2& 79.7& 88.6& 89.3& 92.5& 62.0 &\colorbox{lightgray}{86.9}\\
AD-CLIP~\cite{singha2023ad}&ICCV'23&98.1 &83.6& 91.2& 76.6 &98.1& 93.4 &96.0& 81.4& 86.4& 91.5& 92.1 &64.2& \colorbox{lightgray}{87.7}\\
PADCLIP~\cite{lai2023padclip}&ICCV'23&96.7& \textbf{88.8}& 87.0& 82.8& 97.1 &93.0& 91.3& 83.0& 95.5& 91.8& 91.5& 63.0 &\colorbox{lightgray}{88.5}\\
UniMoS~\cite{li2024split}&CVPR'23&97.7& 88.2 &90.1& 74.6 &96.8& 95.8& 92.4 &84.1& 90.8&89.0 &91.8&65.3 &\colorbox{lightgray}{88.1}\\
DAMP~\cite{du2024domain}&CVPR'23&97.3& 91.6& 89.1& 76.4 &97.5& 94.0& 92.3& 84.5& 91.2& 88.1& 91.2& 67.0 &\colorbox{lightgray}{88.4}\\
\hline
\textbf{CLIP-Div (ResNet-101)}&-&97.2&77.5&91.2&75.7&95.7&69.1&91.8&64.5&78.0&82.2&92.3&54.6&\colorbox{lightgray}{80.8}\\
\textbf{CLIP-Div (ViT-B)}&-&\textbf{98.7}&84.5&92.6&77.0&97.2&89.7&\textbf{97.5}&72.0&85.0&\textbf{95.8}&96.1&66.9&\colorbox{lightgray}{87.8}\\
\textbf{CLIP-Div (ViT-L)}&-&\textbf{98.7}&85.1&\textbf{92.8}&80.2&96.9&88.7&96.8&77.7&86.5&95.6&\textbf{96.8}&\textbf{68.3}&\colorbox{lightgray}{\textbf{88.7}}\\
\hline
\end{tabular}
\label{tab:VisDA1}
\vspace{-10pt}
\end{table*}

\begin{table*}[t]
\caption{Comparison with SoTA methods on DomainNet.}
\centering
\resizebox{\linewidth}{!}{ 
\begin{tabular}{|c||lllllll||c||lllllll||c||lllllll|}
\hline
MCD~\cite{SaitoWUH18}                & clp& inf& pnt& qdr& rel& skt& Avg &SWD~\cite{LeeBBU19}& clp& inf& pnt& qdr& rel& skt& Avg&BNM~\cite{CuiWZLH020} & clp& inf& pnt& qdr& rel& skt& Avg  \\
\hline
\hline
clp                      &  - &15.4 &25.5& 3.3& 44.6& 31.2 &24.0  &clp&  -& 14.7& 31.9& 10.1& 45.3& 36.5& 27.7&clp                      &  -& 12.1 &33.1& 6.2& 50.8 &40.2& 28.5  \\
inf                     &  24.1& -& 24.0 &1.6 &35.2& 19.7& 20.9 &inf &   22.9& -& 24.2& 2.5& 33.2& 21.3& 20.0&inf                     &   26.6 &- &28.5& 2.4 &38.5& 18.1& 22.8\\
pnt                      & 31.1& 14.8& -& 1.7& 48.1& 22.8& 23.7& pnt &  33.6 &15.3 &- &4.4 &46.1& 30.7 &26.0&pnt                      &  39.9& 12.2& - &3.4& 54.5& 36.2 &29.2\\
qdr                      &  8.5& 2.1 &4.6& -& 7.9& 7.1& 6.0& qdr &  15.5 &2.2& 6.4& -& 11.1& 10.2 &9.1&qdr                      &  17.8 &1.0 &3.6& - &9.2& 8.3& 8.0\\
rel                      &  39.4& 17.8& 41.2 &1.5& -& 25.2& 25.0&real& 41.2& 18.1& 44.2& 4.6& -& 31.6& 27.9 &rel                      & 48.6 &13.2& 49.7& 3.6& - &33.9& 29.8 \\
skt                     &   37.3& 12.6& 27.2 &4.1 &34.5& - &23.1& skt & 44.2& 15.2&37.3 &10.3& 44.7& -& 30.3 &skt                     &   54.9 &12.8 &42.3 &5.4 &51.3& - &33.3  \\ 
Avg                      &  28.1 &12.5& 24.5& 2.4& 34.1& 21.2 &\colorbox{lightgray}{20.5}& Avg & 31.5& 13.1& 28.8 &6.4& 36.1& 26.1& \colorbox{lightgray}{23.6}&Avg                      & 37.6& 10.3& 31.4& 4.2& 40.9& 27.3 & \colorbox{lightgray}{25.3} \\
\hline
\hline
CGDM~\cite{du2021cross} & clp& inf& pnt& qdr& rel& skt& Avg&SCDA~\cite{0008XLLLQL21} & clp& inf& pnt& qdr& rel& skt& Avg&CLIP-Div (ViT-B) & clp& inf& pnt& qdr& rel& skt& Avg \\
\hline
\hline
clp& -& 16.9& 35.3& 10.8& 53.5 &36.9& 30.7 & clp& -& 18.6& 39.3& 5.1& 55.0& 44.1& 32.4&clp&-&39.4&66.2&23.8&80.4&64.4&54.8\\
inf &27.8 &- &28.2& 4.4 &48.2& 22.5& 26.2& inf &29.6& - &34.0 &1.4 &46.3& 25.4& 27.3&inf&73.9&-&65.7&13.6&80.1&63.8&59.4 \\ 
pnt& 37.7& 14.5& - &4.6& 59.4& 33.5& 30.0  &pnt& 44.1& 19.0& -& 2.6& 56.2& 42.0& 32.8&pnt&74.5&39.6&-&16.6&81.4&65.0&55.4 \\
qdr& 14.9 &1.5& 6.2& - &10.9& 10.2& 8.7  &qdr& 30.0 &4.9 &15.0 &- &25.4 &19.8 &19.0&qdr&70.9&15.1&38.2&-&74.2&51.9&50.1\\
rel &49.4& 20.8& 47.2& 4.8& - &38.2& 32.0 & rel &54.0 &22.5& 51.9 &2.3& - &42.5& 34.6&rel&74.3&39.3&66.3&18.2&-&63.4&52.3\\
skt& 50.1&16.5& 43.7& 11.1& 55.6& - &35.4 & skt& 55.6 &18.5 &44.7 &6.4 &53.2 &- &35.7&skt&75.6&39.0&67.3&19.0&78.3&-&55.8\\ 
Avg& 36.0& 14.0& 32.1& 7.1 &45.5& 28.3& \colorbox{lightgray}{27.2} &Avg& 42.6& 16.7& 37.0& 3.6 &47.2& 34.8& \colorbox{lightgray}{30.3}&Avg&73.8&34.5&60.7&18.2&78.9&61.7&\colorbox{lightgray}{\textbf{54.6}}\\
\hline
\end{tabular}}
\label{tab:domainnet}
\vspace{-12pt}
\end{table*}

\noindent \textbf{Results on DomainNet.} Tab.~\ref{tab:domainnet} presents a performance comparison on the challenging DomainNet dataset utilizing ResNet-50. Our CLIP-Div (ViT-B) achieves a notable average accuracy of \textbf{54.6}\%, surpassing baseline methods such as CGDM~\cite{du2021cross} and SCDA~\cite{0008XLLLQL21}. Our CLIP-Div (ViT-B) exhibits a significant performance improvement over the baseline methods across \textbf{all transfer tasks}, achieving an average accuracy enhancement of \textbf{24.3}\%. These results demonstrate the effectiveness of CLIP-Div for domain alignment.


\begin{figure*}[t]
    \centering
    \includegraphics[width=\linewidth]{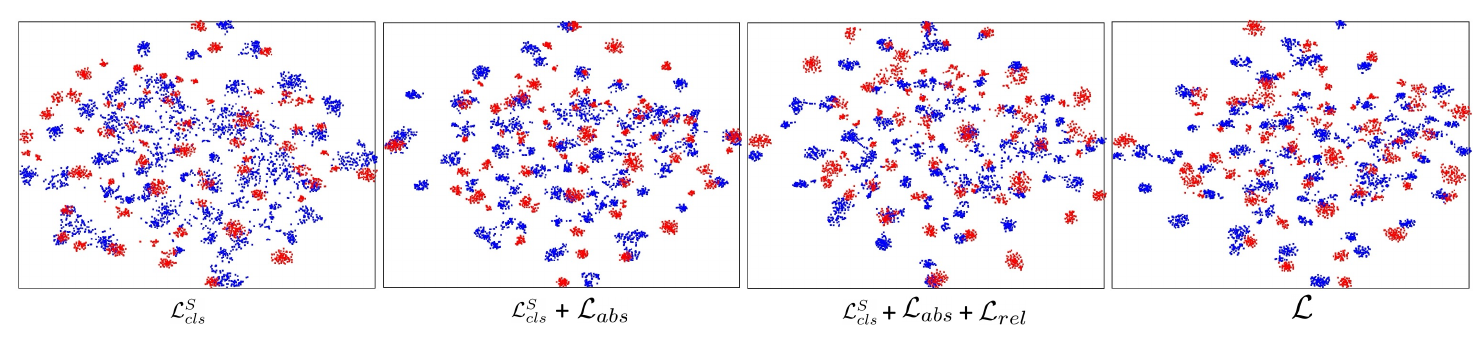}
      \vspace{-20pt}
    \caption{The t-SNE visualization with the proposed losses on the task A $\to$ P.}
    \label{fig:t-sne}
    \vspace{-17pt}
\end{figure*}

\begin{table*}[t]
\caption{Effect of each component of CLIP-Div (ViT-B) on Office-Home.}
\centering
\resizebox{0.9\linewidth}{!}{ 
\begin{tabular}{|cccc||lllllllllllll|}
\hline
$\mathcal{L}_{cls}^{S}$&$\mathcal{L}_{abs}$&$\mathcal{L}_{rel}$&$\mathcal{L}_{pl}$&A$\to$ C& A$\to$ P & A $\to$ R & C $\to$ A &C $\to$ P &C $\to$ R &P$\to$ A& P$\to$ C & P$ \to$ R & R$ \to$ A &R$ \to$ C &R$ \to$ P& Avg     \\
\hline
\textcolor{red}{\ding{51}}&&&&50.8&67.9&75.5&54.5&64.6&66.2&54.0&45.6&75.0&66.1&52.9&79.6&\colorbox{lightgray}{62.7}\\
\textcolor{red}{\ding{51}}&\textcolor{red}{\ding{51}}&&&67.6&86.3&87.1&81.0&87.1&87.4&79.8&69.2&88.5&81.3&70.3&88.0&\colorbox{lightgray}{81.1}\\
\textcolor{red}{\ding{51}}& &\textcolor{red}{\ding{51}}&&68.8&88.0&88.3&81.9&89.4&88.4&80.8&69.5&89.0&81.1& 70.5&89.1&\colorbox{lightgray}{82.0}\\
\textcolor{red}{\ding{51}}& &&\textcolor{red}{\ding{51}}&56.9&78.3&78.7&58.0&76.3&72.3&58.5&53.4&78.2&66.1&57.9&82.8&\colorbox{lightgray}{68.1}\\
\textcolor{red}{\ding{51}}&\textcolor{red}{\ding{51}} &\textcolor{red}{\ding{51}}&&69.2&88.5&88.7&82.7&89.1&88.6&81.3&70.6&89.3&82.4&70.8&89.2&\colorbox{lightgray}{82.5}\\
\textcolor{red}{\ding{51}} & \textcolor{red}{\ding{51}}& \textcolor{red}{\ding{51}}&\textcolor{red}{\ding{51}}&71.4&89.5&89.4&83.0&89.9&89.0&81.5&73.0&89.4&81.8&73.9&90.2&\colorbox{lightgray}{83.5}\\
\hline
\end{tabular}}
\label{tab:ablation_study}
\vspace{-10pt}
\end{table*}


\begin{table*}[t]
\caption{Sensitivity of $\lambda_{abs}$,  $\lambda_{abs}$, and $\lambda_{pl}$ of CLIP-Div (ViT-B) on Office-Home.}
\centering
\resizebox{0.9\linewidth}{!}{ 
\begin{tabular}{|c||lllll||c||lllll||c||lllll|}
\hline
$\lambda_{abs}$&A&C&P&R&Avg &$\lambda_{rel}$&A&C&P&R&Avg&$\lambda_{pl}$&A&C&P&R&Avg\\
\hline
0.0&58.2&49.8&70.7&72.2&\colorbox{lightgray}{62.7}&0.0&80.7&69.0&87.1&87.7&\colorbox{lightgray}{81.1}&0.0&82.1&70.2&88.9&88.9&\colorbox{lightgray}{82.5}\\
1.0&74.4&61.8&81.7&81.3&\colorbox{lightgray}{74.8}&0.1&81.1&69.4&87.5&88.4&\colorbox{lightgray}{81.6}&0.1&\textbf{82.1}&\textbf{72.8}&\textbf{89.9}&\textbf{89.3}&\colorbox{lightgray}{\textbf{83.5}}\\
5.0&\textbf{80.8}&68.4&86.7&87.0&\colorbox{lightgray}{80.7}&0.5&81.9&70.0&88.8&\textbf{88.9}&\colorbox{lightgray}{82.4}&0.2&81.5&73.0&89.8&89.0&\colorbox{lightgray}{83.3}\\
10.&80.7&\textbf{69.0}&\textbf{87.1}&\textbf{87.7}&\colorbox{lightgray}{\textbf{81.1}}&1.0&\textbf{82.1}&\textbf{70.2}&\textbf{88.9}&\textbf{88.9}&\colorbox{lightgray}{\textbf{82.5}}&0.5&77.8&69.9&88.6&87.5&\colorbox{lightgray}{81.0}\\
20.&80.6&68.7&87.0&87.3&\colorbox{lightgray}{80.9}&2.0&81.5&70.0&\textbf{88.9}&88.5&\colorbox{lightgray}{82.3}&1.0&76.6&68.2&87.6&86.2&\colorbox{lightgray}{79.6}\\
\hline
\end{tabular}}
\label{tab:sensitivity}
\vspace{-15pt}
\end{table*}

\begin{table*}[t]
 \caption{Comparison of CLIP-Div with existing methods, validated on Office-Home.}
\centering
\resizebox{0.7\linewidth}{!}{ 
\begin{tabular}{|c||l|l|l|l|l|}
\hline
Method&Language guidance&Prompt learning& Fine-tune CLIP&Inference model&Avg\\
\hline
ResNet-50&\textcolor{green}{\ding{55}}&\textcolor{green}{\ding{55}}&\textcolor{green}{\ding{55}}&ResNet-50& \colorbox{lightgray}{59.4} \\
FixBi~\cite{NaJCH21}&\textcolor{green}{\ding{55}}&\textcolor{green}{\ding{55}}&\textcolor{green}{\ding{55}}&ResNet-50&\colorbox{lightgray}{72.7}\\
kSHOT~\cite{sun2022prior}&\textcolor{green}{\ding{55}}&\textcolor{green}{\ding{55}}&\textcolor{green}{\ding{55}}&ResNet-50&\colorbox{lightgray}{73.9}\\
\hline
AD-CLIP~\cite{singha2023ad}&\textcolor{green}{\ding{55}}&\textcolor{red}{\ding{51}}&\textcolor{green}{\ding{55}}&CLIP (ResNet-50)&\colorbox{lightgray}{75.9}\\
DAMP~\cite{du2024domain}&\textcolor{green}{\ding{55}}&\textcolor{red}{\ding{51}}&\textcolor{green}{\ding{55}}&CLIP (ResNet-50)&\colorbox{lightgray}{78.2}\\
PADCLIP~\cite{lai2023padclip}&\textcolor{green}{\ding{55}}&\textcolor{green}{\ding{55}}&\textcolor{red}{\ding{51}}&CLIP (ResNet-50)&\colorbox{lightgray}{76.6}\\
\hline
CLIP-Div (ResNet-50)&\textcolor{red}{\ding{51}}&\textcolor{green}{\ding{55}}&\textcolor{green}{\ding{55}}&ResNet-50&\colorbox{lightgray}{73.9}\\
CLIP-Div (ViT-B)&\textcolor{red}{\ding{51}}&\textcolor{green}{\ding{55}}&\textcolor{green}{\ding{55}}&ResNet-50&\colorbox{lightgray}{83.5}\\
CLIP-Div (ViT-L)&\textcolor{red}{\ding{51}}&\textcolor{green}{\ding{55}}&\textcolor{green}{\ding{55}}&ResNet-50&\colorbox{lightgray}{\textbf{88.5}}\\
\hline
\end{tabular}}
\label{tab:backbone2}
\vspace{-15pt}
\end{table*}


\subsection{Analysis and Discussion}
In this section, we conduct a comprehensive ablation studies on the Office-Home dataset using CLIP-Div (ViT-B).

\textbf{Effect of each component of CLIP-Div.} In evaluating the individual contributions of each component within CLIP-Div, we conduct an ablation study on the Office-Home dataset, as presented in Tab.~\ref{tab:ablation_study}. Relative to ResNet-50, the incorporation of absolute divergence (\textbf{81.1}\%), relative divergence (\textbf{82.0}\%), and language-guided pseudo-labeling strategy (\textbf{68.1}\%) result in significant classification accuracy improvements, with average boosts of \textbf{18.4}\%, \textbf{19.3}\%, and \textbf{5.4}\%, respectively. These substantial enhancements underscore the effectiveness of employing language-guided divergence measurement and pseudo-labeling strategy for UDA. Furthermore, through a gradual integration of the three components, our experimental results demonstrate performance improvements of \textbf{18.4}\%, \textbf{1.4}\%, and \textbf{1.0}\%, respectively. These observations affirm the cumulative effectiveness of our proposed method, highlighting its ability to synergistically leverage these components for enhancing the UDA model's performance. Additionally, we visualize the features learned by each component of CLIP-Div (ViT-B) on the task A $\to$ P via the t-SNE ~\cite{DonahueJVHZTD14}. The visualization in Fig. \ref{fig:t-sne} further demonstrates the effectiveness of each component of CLIP-Div to align domains.

\noindent\textbf{Sensitivity of hyper-parameters.} In investigating the impact of hyper-parameters on model performance, an extensive analysis is conducted using the Office-Home dataset. 
Three key hyper-parameters, namely $\lambda_{abs}$, $\lambda_{rel}$, and $\lambda_{pl}$, are considered, as detailed in Tab.~\ref{tab:sensitivity}. The results with these hyper-parameters reveal that our proposed loss design proves more effective in mitigating domain divergence and enhancing model generalization with language guidance. Ultimately, to achieve superior performance in UDA, we set $\lambda_{abs} = 10$, $\lambda_{rel} = 1$, and $\lambda_{pl} = 0.1$ to balance each term within our proposed CLIP-Div. 

\noindent\textbf{Importance of vision backbone of CLIP.}  Our approach leverages CLIP as a critical component, serving as a crucial tool to facilitate the training of our UDA model. Specifically, CLIP plays a pivotal role in aligning domains and calibrating the target pseudo labels to enhance the generalization ability of our model. To ascertain the significance of the vision backbone of CLIP, we conduct a comprehensive ablation study, the results of which are presented in Tab.~\ref{tab:backbone}. The findings show
that our CLIP-Div consistently outperforms CLIP across various vision backbones. Specifically, CLIP-Div exhibits superior performance compared to CLIP with ResNet-50, ResNet-101, ViT-B, and ViT-L vision backbones, achieving average accuracy improvements of \textbf{7.8}\%, \textbf{6.2}\%, \textbf{4.0}\%, and \textbf{3.0}\%, respectively, on the Office-Home dataset. Moreover, \textit{as the capacity of CLIP's visual backbone expands, CLIP can offer a more reliable domain-agnostic distribution, thus serving as a superior bridge for domain alignment.}

\begin{table}[t]
 \caption{Ablation results about vision backbone of CLIP on Office-Home. }
\centering
\begin{tabular}{|c||lllll|}
\hline
Method&A&C&P&R&Avg \\
\hline
CLIP (ResNet-50)&65.3 &48.8&75.2&75.2&\colorbox{lightgray}{66.1}\\
CLIP-Div (ResNet-50)&73.3 &58.6 &81.0 &82.5&\colorbox{lightgray}{73.9}$_{\textcolor{red}{+7.8}}$\\
\hline
CLIP (ResNet-101)&70.3 &56.1&78.4&78.2&\colorbox{lightgray}{70.8}\\
CLIP-Div (ResNet-101)&76.2&62.7&83.8&85.2&\colorbox{lightgray}{77.0}$_{\textcolor{red}{+6.2}}$\\
\hline
CLIP (ViT-B)& 78.4&66.7&86.6&86.3&\colorbox{lightgray}{79.5}\\
CLIP-Div (ViT-B)&82.1&72.8&89.9&89.3&\colorbox{lightgray}{83.5}$_{\textcolor{red}{+4.0}}$\\
\hline
CLIP (ViT-L)&83.6&75.1&92.4&90.9&\colorbox{lightgray}{85.5}\\
CLIP-Div (ViT-L)&87.1&79.9&93.7&93.3&\colorbox{lightgray}{88.5}$_{\textcolor{red}{+3.0}}$\\
\hline
\end{tabular}
\label{tab:backbone}
\vspace{-10pt}
\end{table}


\noindent\textbf{CLIP-Div vs. fine-tuning or prompting CLIP\cite{singha2023ad, lai2023padclip}.} In Tab.~\ref{tab:backbone2}, 
the comparative analysis reveals several key observations: \textbf{1)} Our CLIP-Div, leveraging CLIP (ResNet-50) as language guidance, achieves performance on par with ResNet-based methods like kSHOT~\cite{sun2022prior}, albeit falling short compared to CLIP-based methods like AD-CLIP~\cite{singha2023ad} and PADCLIP~\cite{lai2023padclip}. \textbf{2)} This performance gap can be attributed to our approach, involving designated prompts and fixed visual and text encoders of CLIP to derive \textit{a domain-agnostic distribution for bridging domains}. Conversely, SoTA CLIP-based methods employ prompt learning (e.g., AD-CLIP~\cite{singha2023ad}) or fine-tuning the visual encoder of CLIP (e.g., PADCLIP~\cite{lai2023padclip}), potentially incurring high computation costs. \textbf{3)} Our proposed CLIP-Div with CLIP (ViT-B or ViT-L) as language guidance achieves \textbf{83.5}\% (or \textbf{88.5}\%) accuracy, largely outperforming DAMP~\cite{du2024domain} (78.2\%). The observed performance variance in CLIP-Div, with language guidance from different visual encoders of CLIP, underscores the significant impact of the domain-agnostic distribution learned from CLIP on the effectiveness of bridging domains and subsequent domain alignment. Furthermore, this finding substantiates the rationale behind our intuition:\textit{\textbf{ reducing the gap between the two distributions and the domain-agnostic distribution – as a crucial bridge – can facilitate the domain alignment}}. \textbf{4)} In the inference process, our proposed CLIP-Div with ResNet-50 backbone surpasses all the ResNet-based and CLIP-based methods under the language guidance from CLIP (ViT-B or ViT-L).

\noindent\textbf{Computational complexity.} To illustrate the efficiency of our CLIP-Div, we present the training parameters, and training and inference times alongside Swin-based PMTrans~\cite{zhu2023patch} in Tab.~\ref{tab:para}. The results robustly underscore the effectiveness and efficiency of ResNet-based CLIP-Div (ViT-L) for aligning domains in comparison to Swin-based PMTrans~\cite{zhu2023patch}.

\begin{table*}[t]
\caption{Quantitative results on the task A $\to$ P (The time corresponds to one epoch). }
\centering
\begin{tabular}{|c||llll|}
\hline
Method&Training parameters (M) &Training time (s)& Inference time (s)&Accuracy\\
\hline
PMTrans~\cite{zhu2023patch}&88.91&126.2&29.4&91.6\\
CLIP-Div (ViT-B)&\textbf{24.59} &\textbf{66.6}&27.8&89.5\\
CLIP-Div (ViT-L)&\textbf{24.59}&82.1&\textbf{26.6}&\textbf{93.7}\\
\hline
\end{tabular}
\label{tab:para}
\vspace{-15pt}
\end{table*}

\begin{table*}[t]
\caption{Quantitative results on the task A $\to$ P (The time corresponds to one epoch). }
\centering
\resizebox{0.9\linewidth}{!}{ 
\begin{tabular}{|c||lllllllllllll|}
\hline
Method &A$\to$ C& A$\to$ P & A $\to$ R & C $\to$ A &C $\to$ P &C $\to$ R &P$\to$ A& P$\to$ C & P$ \to$ R & R$ \to$ A &R$ \to$ C &R$ \to$ P& Avg     \\
\hline
ResNet-50~\cite{he2016deep} & 44.9& 66.3 &74.3 &51.8& 61.9 &63.6& 52.4 &39.1& 71.2& 63.8& 45.9& 77.2 &\colorbox{lightgray}{59.4}   \\
G-SFDA~\cite{yang2021generalized}&57.9 &78.6 &81.0& 66.7 &77.2& 77.2& 65.6& 56.0& 82.2& 72.0 &57.8 &83.4& \colorbox{lightgray}{71.3}   \\
SHOT~\cite{liang2020we}&57.1 &78.1 &81.5& 68.0 &78.2& 78.1& 67.4& 54.9& 82.2& 73.3 &58.8 &84.3& \colorbox{lightgray}{71.8}   \\
NRC~\cite{yang2021exploiting}&57.7 &80.3 &82.0& 68.1 &79.8& 78.6& 65.3& 56.4& 83.0& 71.0 &58.6 &85.6&\colorbox{lightgray} {72.2}   \\
$A^{2}$Net~\cite{xia2021adaptive}&58.4 &79.0 &82.4& 67.5 &79.3& 78.9& 68.0& 56.2& 82.9& 74.1 &60.5 &85.0& \colorbox{lightgray}{72.8}   \\
SFDA-DE~\cite{ding2022source}&59.7 &79.5 &82.4& 69.7 &78.6& 79.2& 66.1& 57.2& 82.6& 73.9 &60.8 &85.5& \colorbox{lightgray}{72.9}   \\
\textbf{CLIP-Div (ViT-B)}&\textbf{68.9}&\textbf{88.6}&\textbf{88.3}&\textbf{81.5}&\textbf{88.7}&\textbf{88.6}&\textbf{80.7}&\textbf{72.0}&\textbf{88.4}&\textbf{81.4}&\textbf{71.5}&\textbf{89.3}&\colorbox{lightgray}{\textbf{82.3}}\\
\hline
\end{tabular}}
\label{tab:officehome_sourcefree}
\vspace{-10pt}
\end{table*}

\noindent\textbf{Relation with knowledge distillation (KD).} \textbf{1)} KD aims to transfer knowledge from a large teacher model to a compact student model by minimizing the discrepancy between their representations~\cite{gou2021knowledge, wang2021knowledge, hinton2015distilling, zhu2023good}. Differently, our CLIP-Div is a type of UDA approach for measuring the domain divergence with language guidance. Specifically,  
it aims to reduce the gap between the two distributions and the domain-agnostic distribution and then facilitate the domain alignment. 
Therefore, we propose two different prompts and divergence measurement losses to measure the domain divergence between the domain-specific distributions and the domain-agnostic distribution.
\textbf{2)} Similarly, the domain-agnostic distribution from CLIP serves as the teacher, and the domain-specific distribution from the UDA model serves as the student. These language-guided divergence losses enable domain-specific distributions to mimic the domain-agnostic distribution, aligning our approach with knowledge distillation principles. Thus, the reliability of the domain-agnostic distribution derived from CLIP is crucial for bridging domains, as evidenced by the ablation study highlighting the importance of CLIP's vision backbone.

\noindent\textbf{Impact of using language guidance for pseudo-labeling.} 
To assess the effectiveness of the proposed language-guided pseudo-labeling strategy, we conduct a comparative analysis with a previous pseudo-labeling method~\cite{liang2020we} (as depicted in Tab.~\ref{tab:text_pseudo}), revealing its superior performance by surpassing the baseline by \textbf{3.7}\% in average accuracy. Remarkably, the language-guided pseudo-labeling strategy achieves accuracy rates of \textbf{60.9}\%, \textbf{56.1}\%, \textbf{79.1}\%, and \textbf{76.4}\% across respective target tasks. These compelling findings underscore the commendable generalization ability of our model on the target data, particularly when trained with language guidance. Additionally, we compare our language-guided pseudo-labeling approach with the adaptive confidence-based pseudo-labeling method~\cite{NaJCH21}, further highlighting the superior performance of our approach.  

\begin{table}[t]
\caption{Ablation results of language-guided pseudo labeling on Office-Home.}
\centering
\resizebox{0.9\linewidth}{!}{

\begin{tabular}{|c||lllll|}
\hline
Method&A&C&P&R&Avg \\
\hline
ResNet-50&58.2&49.8&70.7&72.2&\colorbox{lightgray}{62.7}\\
Pseudo-labeling~\cite{liang2020we}&56.7&51.2&74.1&75.4&\colorbox{lightgray}{64.4}\\
Adaptive confidence~\cite{NaJCH21}&60.5&53.7&\textbf{79.2}&\textbf{76.8}&\colorbox{lightgray}{67.6}\\
Language-guided pseudo-labeling&\textbf{60.9}&\textbf{56.1}&79.1&76.4&\colorbox{lightgray}{\textbf{68.1}}\\
\hline
\end{tabular}}
\label{tab:text_pseudo}
\vspace{-10pt}
\end{table}

\noindent\textbf{Impact of $\textbf{prompt}^{a}$ and $\textbf{prompt}^{avg}$ for $\mathcal{L}_{abs}$ and $ \mathcal{L}_{rel}$.} In this study, we aim to comprehensively explore the intrinsic knowledge of CLIP for UDA. To achieve this, we propose the utilization of both domain-agnostic and domain-specific prompts for $\mathcal{L}_{abs}$ and $\mathcal{L}_{rel}$, respectively. In Tab.~\ref{tab:prompt}, we investigate the impact of employing $\textbf{prompt}^a$ and $\textbf{prompt}^{avg}$ for the proposed losses $\mathcal{L}_{abs}$ and $\mathcal{L}_{rel}$. The results indicate that our proposed CLIP-Div, with $\textbf{prompt}^a$ in $\mathcal{L}_{abs}$ and $\textbf{prompt}^{avg}$ in $\mathcal{L}_{rel}$, achieves the best performance in three target tasks and the highest average performance on the Office-Home dataset. This comparison underscores that utilizing different prompts for two losses facilitates a comprehensive exploration of CLIP for UDA. 

\begin{table}[t]
\caption{Ablation results about $\textbf{prompt}^a$ and $\textbf{prompt}^{avg}$ for $\mathcal{L}_{abs}$ and $ \mathcal{L}_{rel}$ on Office-Home.}
\centering
\resizebox{0.9\linewidth}{!}{ 
\begin{tabular}{|c||lllll|}
\hline
Method&A&C&P&R&Avg \\
\hline
($\textbf{prompt}^a$, $\textbf{prompt}^{avg}$)&\textbf{82.1}& \textbf{72.8} &89.9& \textbf{89.3}&\colorbox{lightgray}{\textbf{83.5}}\\
($\textbf{prompt}^{avg}$, $\textbf{prompt}^{avg}$)&82.0&\textbf{72.8}&\textbf{90.1}&89.2&\colorbox{lightgray}{\textbf{83.5}}\\
($\textbf{prompt}^{avg}$, $\textbf{prompt}^a$)&81.8&72.7&89.6&89.1&\colorbox{lightgray}{83.3}\\
($\textbf{prompt}^a$, $\textbf{prompt}^a$)&81.4&72.1&89.2&89.0&\colorbox{lightgray}{82.7}\\
\hline
\end{tabular}}
\label{tab:prompt}
\vspace{-10pt}
\end{table}

\begin{table}[t]
\caption{Comparison with ViT-based methods on DomainNet.}
\centering
\resizebox{0.9\linewidth}{!}{ 
\begin{tabular}{|c||lllllll|}
\hline
Method&clp&inf&pnt&qdr&rel&skt&Avg \\
\hline
CDTrans~\cite{abs-2109-06165}&59.9&25.3&52.2&19.6&65.9&48.4&\colorbox{lightgray}{45.2}\\
SSRT~\cite{abs-2204-07683}&60.0 &28.2& 53.3 &13.7& 65.3 &50.4& \colorbox{lightgray}{45.2}\\
PMTrans~\cite{zhu2023patch}&67.9&30.7&59.1&27.0&72.8&56.9&\colorbox{lightgray}{52.4}\\
PADCLIP~\cite{lai2023padclip}&75.3&\textbf{54.6}&71.2&\textbf{30.7}&83.6&67.1&\colorbox{lightgray}{\textbf{63.7}}\\
UniMoS~\cite{li2024split}&\textbf{77.1}&55.0&\textbf{71.7}&24.1&\textbf{85.8}&\textbf{68.0}&\colorbox{lightgray}{63.6}\\
CLIP-Div (ViT-B)&73.8 &34.5& 60.7& 18.2 &78.9 &61.7 &\colorbox{lightgray}{54.6} \\
\hline
\end{tabular}}
\label{tab:vit_comp}
\vspace{-10pt}
\end{table}

\noindent\textbf{Comparison with ViT-based methods.} In the comparison with ViT-based~\cite{abs-2109-06165, abs-2204-07683} and Swin-based~\cite{zhu2023patch} methods in Tab.~\ref{tab:vit_comp}, our ResNet-based CLIP-Div (ViT-B) outperforms all ViT-based~\cite{abs-2109-06165, abs-2204-07683} and Swin-based~\cite{zhu2023patch} methods, except for PADCLIP~\cite{lai2023padclip} and UniMoS~\cite{li2024split} on the most challenging dataset. This underscores the effectiveness of our proposed CLIP-Div in learning the domain-agnostic distribution for alleviating the domain divergence effectively.

\noindent\textbf{Applications on source-free domain adaptation.} To evaluate the generalization capability of CLIP-Div, we leverage its application in addressing source-free domain adaptation (SFDA)  challenges. Our experimentation focuses on the Office-Home dataset, where we compare CLIP-Div against SoTA SFDA methodologies, including G-SFDA~\cite{yang2021generalized}, SHOT~\cite{liang2020we}, NRC~\cite{yang2021exploiting}, A$^2$Net~\cite{xia2021adaptive}, and SFDA-DE~\cite{ding2022source}. The results, as presented in Tab.~\ref{tab:officehome_sourcefree}, unequivocally demonstrate that CLIP-Div outperforms all considered SFDA methods across various transfer tasks. Notably, CLIP-Div attains an impressive \textbf{82.3}\% average accuracy, showcasing a remarkable \textbf{9.4}\% accuracy improvement. These findings underscore the robust generalization capacity of our proposed approach in diverse UDA scenarios and affirm the effectiveness of CLIP-Div.


\section{Conclusion and Future Work}
\label{sec:conclusion}
In this paper, we introduced a novel language-guided approach, named CLIP-Div, to harness the semantic richness and zero-shot generalization capability of CLIP to improve the UDA model's performance. Our main contributions include introducing two innovative language-guided divergence measurements, namely absolute and relative divergence, to align the source and target domains. We also present a language-guided pseudo-labeling strategy for calibrating target pseudo-labels, followed by self-training to further improve the UDA model's generalization ability on the target domain. Extensive experimentation has underscored the effectiveness of achieving SoTA performance on four benchmark datasets, surpassing leading prior methods by a substantial margin. 

\noindent\textbf{Limitation}: 
The reliability of the domain-agnostic distribution derived from a fixed CLIP model is critical for bridging the domain gap. The effectiveness of the UDA model varies when changing CLIP's visual encoders from ViT-based models to ResNet-based ones, such as ResNet-50 or ResNet-101.


\noindent\textbf{Future work}: Our future work entails implementing CLIP-Div in source-free UDA and domain generalization settings, incorporating language guidance. Moreover, we plan to investigate the potential of leveraging CLIP to effectively bridge domain gaps within the feature space.

\bibliographystyle{./IEEEtran}

\bibliography{./IEEEabrv,./egbib}

\end{document}


\title{CLIP the Divergence: Language-guided Unsupervised Domain Adaptation}

\author{Jinjing Zhu, Yucheng Chen, Lin Wang
}

\markboth{Journal of \LaTeX\ Class Files,~Vol.~14, No.~8, August~2021}%
{Shell \MakeLowercase{\textit{et al.}}: A Sample Article Using IEEEtran.cls for IEEE Journals}


\maketitle

\begin{abstract}
Due to the lack of space in the main paper, we provide more details of the proposed method and experimental results in the supplementary material. Sec.~\ref{ablation} expounds upon the ablation study and associated discussions. Sec.~\ref{source-free} provides empirical evidence showcasing the generalization prowess of CLIP-Div  in source-free domain adaptation. Sec.~\ref{visualization} visually represents the image features extracted by CLIP-Div. Sec.~\ref{algorithm} introduces the algorithm for the proposed CLIP-Div framework. Lastly, Sec.~\ref{prompt_example} illustrates exemplary prompt instances on the Office-Home dataset.
\end{abstract}

\section{Details of Ablation Study}
\label{ablation}

\subsection{Sensitivity of Hyper-Parameters}

In the investigation of hyper-parameter impact on model performance, a comprehensive analysis is undertaken using the Office-Home dataset. Three pivotal hyper-parameters, denoted as $\lambda_{abs}$, $\lambda_{rel}$, and $\lambda_{pl}$, are systematically examined, as delineated in Tabs.~\ref{tab:abs}, \ref{tab:rel}, and \ref{tab:ps} respectively. Within Tab.~\ref{tab:abs}, we explore the sensitivity of the $\lambda_{abs}$ parameter to enhance the effectiveness of the proposed language-guided absolute loss for quantifying domain divergence on the Office-Home dataset. Notably, employing a parameter value of 10 yields an average accuracy of \textbf{81.1}\% and a \textbf{18.2}\% enhancement compared to the ResNet-50~\cite{he2016deep} method, underscoring the significance of the absolute loss in domain alignment.

Likewise, we conduct experiments to assess the sensitivity of hyper-parameters $\lambda_{rel}$ and $\lambda_{pl}$ on the Office-Home dataset, as presented in Tabs.~\ref{tab:rel} and \ref{tab:ps}. Specifically, The relative loss and language-guided pseudo labeling loss yield an average improvement of \textbf{1.4}\% and \textbf{1.0}\%, respectively, compared to baseline methods. The results underscore that our proposed loss design proves more efficacious in mitigating domain divergence and augmenting model generalization with language guidance. In pursuit of superior performance in UDA, we strategically set $\lambda_{abs} = 10$, $\lambda_{rel} = 1$, and $\lambda_{pl} = 0.1$ to attain a balanced integration of each term within our proposed CLIP-Div method.

\begin{table*}[t!]
\centering
\caption{Sensitivity of $\lambda_{abs}$ evaluated on \textbf{Office-Home}. The best performance is marked as \textbf{bold}.}
\resizebox{0.99\linewidth}{!}{ 
\begin{tabular}{|c|cllllllllllll|}
\hline
$\lambda_{abs}$ &A$\to$ C& A$\to$ P & A $\to$ R & C $\to$ A &C $\to$ P &C $\to$ R &P$\to$ A& P$\to$ C & P$ \to$ R & R$ \to$ A &R$ \to$ C &R$ \to$ P& \colorbox{lightgray}{Avg}     \\
\hline
\hline
0.0&50.8&67.9&75.5&54.5&64.6&66.2&54.0&45.6&75.0&66.1&52.9&79.6&\colorbox{lightgray}{62.7}\\
1.0& 59.8&79.0&81.4&71.2&81.1&79.3&73.8&61.3&83.1&78.1&64.2&85.1&\colorbox{lightgray}{74.8}\\
2.0&63.1&82.1&83.1&75.1&84.1&83.6&77.7&65.2&86.4&80.1&67.2&86.5&\colorbox{lightgray}{77.9}\\
5.0&66.7&85.5&86.4&80.8&86.9&86.4&\textbf{80.3}&68.7&88.3&\textbf{81.3}&69.7&87.8&\colorbox{lightgray}{80.7}\\
10.&\textbf{67.6}&86.3&\textbf{87.1}&\textbf{81.0}&\textbf{87.1}&\textbf{87.4}&79.8&\textbf{69.2}&\textbf{88.5}&\textbf{81.3}&\textbf{70.3}&\textbf{88.0}&\colorbox{lightgray}{\textbf{81.1}}\\
20.&67.2&\textbf{86.4}&86.9&80.7&\textbf{87.1}&86.9&80.1&68.9&88.2&81.0&69.9&87.5&\colorbox{lightgray}{80.9}\\
\hline
\end{tabular}}
\label{tab:abs}
\end{table*}

\begin{table*}[t!]
\centering
\caption{Sensitivity of $\lambda_{rel}$ evaluated on \textbf{Office-Home}. The best performance is marked as \textbf{bold}.}
\resizebox{1\linewidth}{!}{
\begin{tabular}{|c|cllllllllllll|}
\hline
$\lambda_{rel}$ &A$\to$ C& A$\to$ P & A $\to$ R & C $\to$ A &C $\to$ P &C $\to$ R &P$\to$ A& P$\to$ C & P$ \to$ R & R$ \to$ A &R$ \to$ C &R$ \to$ P& Avg    \\
\hline
\hline
0.0&67.6&86.3&87.1&81.0&87.1&87.4&79.8&69.2&88.5&81.3&70.3&88.0&\colorbox{lightgray}{81.1}\\
0.1&68.3&86.8&88.4&81.4&87.4&88.0&80.3&69.8&88.7&81.6&70.1&88.3&\colorbox{lightgray}{81.6}\\
0.2&68.9&87.3&88.6&82.0&88.0&88.2&80.7&70.3&88.9&81.6&70.2&88.7&\colorbox{lightgray}{82.0}\\
0.5&69.0&88.4&\textbf{89.2}&82.6&88.9&\textbf{88.6}&\textbf{81.3}&\textbf{70.7}&89.0&81.8&70.5&89.0&\colorbox{lightgray}{82.4}\\
1.0&69.2&\textbf{88.5}&88.7&\textbf{82.7}&89.1&\textbf{88.6}&\textbf{81.3}&70.6&\textbf{89.3}&\textbf{82.4}&\textbf{70.8}&\textbf{89.2}&\colorbox{lightgray}{\textbf{82.5}}\\
2.0&\textbf{69.7}&88.3&88.4&82.0&\textbf{89.3}&88.4&81.0&70.1&88.8&81.7&70.3&89.0&\colorbox{lightgray}{82.3}\\
\hline
\end{tabular}}
\label{tab:rel}
\end{table*}

\begin{table*}[t!]
\centering
\caption{Sensitivity of $\lambda_{rel}$ evaluated on \textbf{Office-Home}. The best performance is marked as \textbf{bold}.}
\resizebox{1\linewidth}{!}{
\begin{tabular}{|c|cllllllllllll|}
\hline
$\lambda_{pl}$ &A$\to$ C& A$\to$ P & A $\to$ R & C $\to$ A &C $\to$ P &C $\to$ R &P$\to$ A& P$\to$ C & P$ \to$ R & R$ \to$ A &R$ \to$ C &R$ \to$ P& Avg    \\
\hline
\hline
0.0&69.2&88.5&88.7&82.7&89.1&88.6&81.3&70.6&89.3&\textbf{82.4}&70.8&89.2&\colorbox{lightgray}{82.5}\\
0.1&\textbf{71.4}&\textbf{89.5}&\textbf{89.4}&\textbf{83.0}&\textbf{89.9}&89.0&\textbf{81.5}&73.0&\textbf{89.4}&81.8&73.9&\textbf{90.2}&\textbf{83.5}\\
0.2&71.2&89.4&88.7&82.4&\textbf{89.9}&\textbf{89.1}&81.1&\textbf{73.5}&89.2&80.9&\textbf{74.3}&90.0&83.3\\
0.5&69.1&88.0&87.4&78.9&88.8&87.3&76.8&70.3&87.9&77.8&70.3&89.0&81.0\\
1.0&65.2&87.3&86.5&77.0&88.2&85.9&75.8&70.1&86.1&77.0&69.3&87.3&79.6\\
\hline
\end{tabular}}
\label{tab:ps}
\end{table*}

\clearpage

\begin{table*}[t]
\centering
\caption{Effect of vision backbone of CLIP evaluated on \textbf{Office-Home}. The best performance is marked as \textbf{bold}.}
\resizebox{0.99\linewidth}{!}{ 
\begin{tabular}{|c|clllllllllllll|}
\hline
backbone &A$\to$ C& A$\to$ P & A $\to$ R & C $\to$ A &C $\to$ P &C $\to$ R &P$\to$ A& P$\to$ C & P$ \to$ R & R$ \to$ A &R$ \to$ C &R$ \to$ P& Avg     \\
\hline
\hline
ResNet-50& 57.2&80.4&82.9&73.9&80.7&81.1&72.8&58.6&83.5&73.3&59.9&81.9&\colorbox{lightgray}{73.9}\\
ResNet-101&61.3&84.1&85.1&77.1&82.8&85.0&75.6&63.1&85.5&75.9&63.7&84.5&\colorbox{lightgray}{77.0}\\
ViT-B&71.4&89.5&89.4&83.0&89.9&89.0&81.5&73.0&89.4&81.8&73.9&90.2&\colorbox{lightgray}{83.5}\\
ViT-L&\textbf{79.1}&\textbf{93.7}&\textbf{93.3}&\textbf{87.7}&\textbf{94.0}&\textbf{93.2}&\textbf{86.8}&\textbf{80.6}&\textbf{93.4}&\textbf{86.8}&\textbf{80.1}&\textbf{93.4}&\colorbox{lightgray}{\textbf{88.5}}\\
\hline
\end{tabular}}
\label{tab:backbone}
\end{table*}

\subsection{Effect of vision backbones of CLIP}

Given our utilization of CLIP's semantic richness and zero-shot generalization capability to alleviate domain divergence and enhance the UDA model's generalization, achieving an optimal domain-agnostic distribution becomes paramount for effective domain alignment. To evaluate the impact of the vision backbone of CLIP, we conduct an ablation study on the Office-Home dataset, as detailed in Tabs.~\ref{tab:backbone} and \ref{tab:backbone1}. The results consistently reveal the superior performance of CLIP-Div compared to CLIP across various vision backbones. Specifically, CLIP-Div exhibits notable enhancements over CLIP when utilizing ResNet-50, ResNet-101, ViT-B, and ViT-L vision backbones, demonstrating average accuracy improvements of \textbf{7.8}\%, \textbf{6.2}\%, \textbf{4.0}\%, and \textbf{3.0}\%, respectively. These improvements underscore the effectiveness of CLIP-Div in diminishing domain divergence and augmenting the UDA model's generalization ability.

\begin{figure}[t]
    \centering
    \includegraphics[width=1\linewidth]{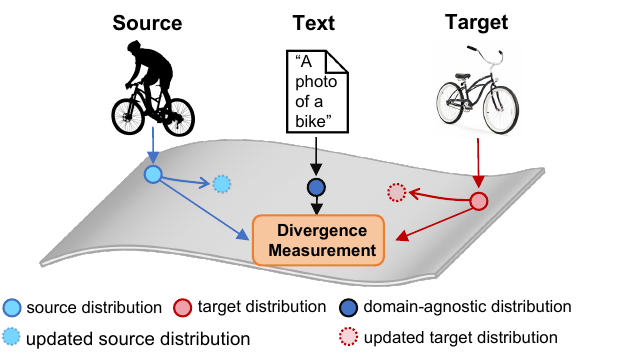}
    \caption{\textbf{The key intuition of our CLIP-Div for measuring the domain divergence}. The domain-agnostic distribution, acquired through CLIP with language guidance, serves as a pivotal bridge between source and target domains through the meticulous design of divergence measurements.}
    \label{fig:intro}
\end{figure}

Moreover, a comparative analysis of the performance of distinct vision backbones and their corresponding CLIP-Div counterparts underscores the critical role of the vision backbone and the associated domain-agnostic distribution. The optimal domain-agnostic distribution is pivotal for superior domain alignment, as evidenced by the observed improvements. Specifically, CLIP with ResNet-50 backbone achieves \textbf{66.1}\% average accuracy and the corresponding CLIP-Div achieves \textbf{73.9}\% average accuracy. And CLIP with ViT-L backbone achieves \textbf{85.5}\% average accuracy and the corresponding CLIP-Div achieves \textbf{88.5}\% average accuracy. The results represent that if the obtained domain-agnostic distribution is optimal, the alignment of two domains with the optimal domain-agnostic distribution will be better. This comparison between the performance of CLIP-Div with ResNet-50 and ViT-L further substantiates the validity of our key intuition illustrated in Fig.~\ref{fig:intro}.

\begin{table}[t!]
\centering
\caption{Ablation results about vision backbone of CLIP on Office-Home.}
\resizebox{\linewidth}{!}{
\begin{tabular}{|c|lllll|}
\hline
Method&A&C&P&R&Avg \\
\hline
\hline
CLIP (ResNet-50)&65.3 &48.8&75.2&75.2&\colorbox{lightgray}{66.1}\\
CLIP-Div&73.3 &58.6 &81.0 &82.5&\colorbox{lightgray}{73.9}$_{\textcolor{red}{+7.8}}$\\
\hline
CLIP (ResNet-101)&70.3 &56.1&78.4&78.2&\colorbox{lightgray}{70.8}\\
CLIP-Div&76.2&62.7&83.8&85.2&\colorbox{lightgray}{77.0}$_{\textcolor{red}{+6.2}}$\\
\hline
CLIP (ViT-B)& 78.4&66.7&86.6&86.3&\colorbox{lightgray}{79.5}\\
CLIP-Div&82.1&72.8&89.9&89.3&\colorbox{lightgray}{83.5}$_{\textcolor{red}{+4.0}}$\\
\hline
CLIP (ViT-L)&83.6&75.1&92.4&90.9&\colorbox{lightgray}{85.5}\\
CLIP-Div&87.1&79.9&93.7&93.3&\colorbox{lightgray}{88.5}$_{\textcolor{red}{+3.0}}$\\
\hline
\end{tabular}}
\label{tab:backbone1}
\end{table}

\begin{table}[t!]
\centering
\caption{Comparison between CLIP and CLIP-Div on office-31.}
\begin{tabular}{|c|llll|}
\hline
Method&A&D&W&Avg \\
\hline
\hline
CLIP (ViT-B)&76.9&79.1&77.0&\colorbox{lightgray}{77.7} \\
CLIP-Div (ViT-B)&81.0&96.2&95.2&\colorbox{lightgray}{90.8}$_{\textcolor{red}{+13.1}}$\\
CLIP(ViT-L)&82.6&84.7&85.7&\colorbox{lightgray}{84.3}\\
CLIP-Div(ViT-L)&\textbf{85.1}&\textbf{97.2}&\textbf{96.1}&\colorbox{lightgray}{\textbf{92.9}}$_{\textcolor{red}{+8.6}}$\\
\hline
\end{tabular}
\label{tab:com_office31}
\end{table}

\begin{table*}[t]
\centering
\caption{Comparison between CLIP and CLIP-Div on VisDA-2017.}
\resizebox{0.99\linewidth}{!}{ 
\begin{tabular}{|c|lllllllllllll|}
\hline
Method &plane& bcycl& bus &car& horse& knife& mcycl& person& plant &sktbrd &train& truck& Avg     \\
\hline
\hline
CLIP (ViT-B)&98.6&89.3&\textbf{94.5}&75.4&98.1&\textbf{89.7}&90.0&74.3&81.6&89.8&93.7&56.9&\colorbox{lightgray}{86.0}\\
CLIP-Div (ViT-B)&98.7&84.5&92.6&77.0&97.2&\textbf{89.7}&\textbf{97.5}&72.0&85.0&95.8&96.1&66.9&\colorbox{lightgray}{87.8}$_{\textcolor{red}{+1.8}}$\\
CLIP (ViT-L)&\textbf{99.7}&\textbf{91.7}&91.3&74.3&\textbf{98.4}&88.8&96.3&59.1&76.3&\textbf{96.2}&\textbf{97.5}&58.8&\colorbox{lightgray}{85.7}\\
CLIP-Div(ViT-L)&98.7&85.1&92.8&\textbf{80.2}&96.9&88.7&96.8&\textbf{77.7}&\textbf{86.5}&95.6&96.8&\textbf{68.3}&\colorbox{lightgray}{\textbf{88.7}}$_{\textcolor{red}{+3.0}}$\\

\hline
\end{tabular}}
\label{tab:com_visda}
\end{table*}

\begin{table}[t!]
\centering
\caption{Comparison between CLIP and CLIP-Div on DomainNet.}
\resizebox{\linewidth}{!}{ 
\begin{tabular}{|c|lllllll|}
\hline
Method&clp&inf&pnt&qdr&rel&skt&Avg \\
\hline
\hline
CLIP (ViT-B)&71.8&\textbf{49.2}&\textbf{65.3}&16.4&\textbf{80.1}&\textbf{64.2}&\textbf{57.8} \\
CLIP-Div (ViT-B)&\textbf{73.8} &34.5 &60.7& \textbf{18.2}& 78.9 &61.7 &54.6$_{\textcolor{blue}{-3.2}}$\\
\hline
\end{tabular}}
\label{tab:com_domainnet}
\end{table}

\begin{table}[t!]
\centering
\caption{Comparison between CLIP and CLIP-Div with ResNet-50 backbone on DomainNet.}
\resizebox{\linewidth}{!}{ 
\begin{tabular}{|c|lllllll|}
\hline
Method&clp&inf&pnt&qdr&rel&skt&Avg \\
\hline
\hline
CLIP (ResNet-50)&55.4&\textbf{38.6}&52.2&7.7&69.4&49.6&45.7\\
CLIP-Div (ViT-B)&\textbf{73.8} &34.5 &\textbf{60.7}& \textbf{18.2}& \textbf{78.9} &\textbf{61.7 }&\textbf{54.6}$_{\textcolor{red}{+8.9}}$\\
\hline
\end{tabular}}
\label{tab:com_domainnet1}
\end{table}


\subsection{Comparison with CLIP}

To validate the effectiveness of our proposed CLIP-Div, a comparative analysis is conducted between CLIP-Div and CLIP across four diverse datasets, namely Office-Home, Office-31, VisDA-2017, and DomainNet, as detailed in Tabs.~\ref{tab:backbone1}, \ref{tab:com_office31}, \ref{tab:com_visda}, and \ref{tab:com_domainnet} respectively.

Specifically, CLIP-Div, when equipped with ViT-B and ViT-L vision backbones, achieves a performance boost of \textbf{4.0}\% and \textbf{3.0}\% in average accuracy on the Office-Home dataset. Furthermore, on the Office-31 dataset, CLIP-Div with ViT-B and ViT-L vision backbones exhibits an average accuracy enhancement of \textbf{13.1}\% and \textbf{8.6}\% respectively. The VisDA-2017 dataset similarly witnesses a performance improvement of \textbf{1.8}\% and \textbf{3.0}\% in average accuracy when utilizing CLIP-Div with ViT-B and ViT-L vision backbones. These observed accuracy enhancements underscore the effectiveness of CLIP-Div in domain alignment and its capacity to enhance the UDA model's generalization on the target domain.

However, on the challenging DomainNet dataset, CLIP-Div performs less favorably than CLIP. This discrepancy may be attributed to the limitation of our CLIP-Div (ViT-B) with a \textbf{ResNet-50} backbone, which does not match the performance of CLIP with ViT-B. To further elucidate, a direct comparison between CLIP-Div (ViT-B) and CLIP (ResNet-50), both equipped with the same \textbf{ResNet-50 backbone}, is presented in Tab.~\ref{tab:com_domainnet1}. Notably, CLIP-Div (ViT-B) achieves a performance boost of \textbf{5.0}\%, reaching an average accuracy of \textbf{54.6}\%, thus emphasizing the effectiveness of our proposed CLIP-Div in addressing UDA challenges. 

In this study, we employ the domain-agnostic distribution acquired from CLIP to mitigate the domain gap between source and target domains. The selection of an optimal domain-agnostic distribution learned from CLIP, particularly with ViT-based vision backbones rather than ResNet-based counterparts, significantly influences the alignment process between the source and target domains.

\clearpage

\subsection{Impact of Using Language Guidance for Pseudo-labeling}
To evaluate the effectiveness of the proposed language-guided pseudo labeling strategy, a comparative analysis with a prior pseudo labeling method (as presented in Tab.~\ref{tab:pseudo}) demonstrates its superior performance, surpassing the baseline by \textbf{3.7}\% in average accuracy. Notably, the language-guided pseudo labeling strategy achieves accuracy rates of \textbf{60.9}\%, \textbf{56.1}\%, \textbf{79.1}\%, and \textbf{76.4}\% across respective target tasks, indicating a substantial performance enhancement. These compelling results underscore the commendable generalization ability of our model on the target data, especially when trained with language guidance.

\subsection{Effect of Batch Size}

To assess the sensitivity of CLIP-Div to batch size variations, we conduct experiments on the Office-Home dataset for UDA tasks. The results, as presented in Tab.~\ref{tab:batchsize}, illustrate the impact of different batch sizes on performance. Our ablation study indicates that the optimal performance for UDA tasks on the Office-Home dataset is achieved with a batch size of 16. Specifically, CLIP-Div with a batch size of 16 attains the highest performance, achieving an average accuracy of \textbf{83.9}\%, outperforming CLIP-Div with other batch sizes. Importantly, the results suggest that batch size variations do not significantly affect the performance of CLIP-Div, affirming its capability to effectively align domains and enhance the generalization of the UDA model on the target domain.

\begin{table*}[t]
\centering
\caption{Effect of each component of CLIP-Div. The best performance is marked as \textbf{bold}. }
\resizebox{0.99\linewidth}{!}{ 
\begin{tabular}{|c|lllllllllllll|}
\hline
Method&A$\to$ C& A$\to$ P & A $\to$ R & C $\to$ A &C $\to$ P &C $\to$ R &P$\to$ A& P$\to$ C & P$ \to$ R & R$ \to$ A &R$ \to$ C &R$ \to$ P& Avg     \\
\hline
\hline
ResNet-50&50.8&67.9&75.5&54.5&64.6&66.2&54.0&45.6&75.0&\textbf{66.1}&52.9&79.6&\colorbox{lightgray}{62.7}\\
Pseudo-labeling&53.3&73.0&78.2&53.9&68.8 &69.9&51.9&45.5&78.0&64.4&54.9&80.6&\colorbox{lightgray}{64.4}\\
Language-guided pseudo labeling&\textbf{56.9}&\textbf{78.3}&\textbf{78.7}&\textbf{58.0}&\textbf{76.3}&\textbf{72.3}&\textbf{58.5}&\textbf{53.4}&\textbf{78.2}&\textbf{66.1}&\textbf{57.9}&\textbf{82.8}&\colorbox{lightgray}{\textbf{68.1}}\\
\hline
\end{tabular}}
\label{tab:pseudo}
\end{table*}
\begin{table*}[t!]
\centering
\caption{Sensitivity of batch size evaluated on \textbf{Office-Home}. The best performance is marked as \textbf{bold}.}
\resizebox{0.99\linewidth}{!}{ 
\begin{tabular}{|c|cllllllllllll|}
\hline
bs &A$\to$ C& A$\to$ P & A $\to$ R & C $\to$ A &C $\to$ P &C $\to$ R &P$\to$ A& P$\to$ C & P$ \to$ R & R$ \to$ A &R$ \to$ C &R$ \to$ P& \colorbox{lightgray}{Avg}     \\
\hline
\hline
8&\textbf{72.8}&89.1&\textbf{89.6}&82.4&89.7&89.1&\textbf{81.9}&\textbf{75.4}&\textbf{89.5}&82.0&\textbf{74.6}&90.0&83.8\\
16&\textbf{72.8}&\textbf{89.9}&89.1&\textbf{83.4}&\textbf{90.4}&\textbf{89.2}&81.8&74.1&89.4&\textbf{82.4}&\textbf{74.6}&\textbf{90.1}&\textbf{83.9}\\
32&71.4&89.5&89.4&83.0&89.9&89.0&81.5&73.0&89.4&81.8&73.9&90.2&83.5\\
64&69.9&89.3&89.1&83.1&\textbf{90.4}&88.9&80.7&71.6&89.2&81.3&71.9&87.8&82.8\\
\hline
\end{tabular}}
\label{tab:batchsize}
\end{table*}

\clearpage

\subsection{Comparison with ViT-based Methods}

To assess the effectiveness of our proposed CLIP-Div, we conduct a comparative analysis against various ViT-based methods on four datasets. Please note that despite unfair for us, we conduct comparative evaluations of ResNet-based CLIP-Div against ViT-based and Swin-based methodologies across four datasets. Our approach, CLIP-Div, employs ResNet backbone for performance assessment, whereas ViT-based and Swin-based methods utilize ViT and Swin backbones, respectively. If our proposed CLIP-Div, utilizing a ResNet backbone, surpasses the performance of ViT-based and Swin-based methods, it will serve as additional evidence for the effectiveness of our approach in addressing UDA challenges.

The considered ViT-based methods encompass ViT~\cite{DBLP:conf/iclr/DosovitskiyB0WZ21}, DAPL~\cite{ge2022domain}, TVT~\cite{abs-2108-05988}, CDTrans~\cite{abs-2109-06165}, SSRT~\cite{abs-2204-07683}, PMTrans~\cite{zhu2023patch}, and PADCLIP~\cite{lai2023padclip}. The comparative results on the four datasets are presented in Tabs.~\ref{tab:officehome_vit}, \ref{tab:office31_vit}, \ref{tab:VisDA_vit}, and \ref{tab:domainnet_vit}. Note that CDTrans uses DeiT and PMTrans leverages Swin transformer.

In Tab.~\ref{tab:officehome_vit}, the results on the Office-Home dataset demonstrate that CLIP-Div outperforms all ViT-based methods except PMTrans~\cite{zhu2023patch}, achieving comparable performance with SoTA PMTrans~\cite{zhu2023patch}. Notably, CLIP-Div (ViT-L) attains the best performance in six transfer tasks, validating its effectiveness in addressing UDA challenges.

However, Tabs.~\ref{tab:office31_vit} and \ref{tab:VisDA_vit} reveal sub-optimal performance on the Office-31 and VisDA-2017 datasets when compared to ViT-based methods. This discrepancy arises from CLIP-Div's use of ResNet as the feature extractor, in contrast to the ViT-based design of the other methods.

In the challenging DomainNet dataset, CLIP-Div surpasses all ViT-based methods except PADCLIP~\cite{lai2023padclip}, as presented in Tab.~\ref{tab:domainnet_vit}. In summary, our proposed CLIP-Div, based on ResNet, achieves comparable performance with SoTA ViT-based and Swin-based methods. This underscores the effectiveness of CLIP-Div in domain alignment and enhancing the UDA model's generalization ability, particularly with the incorporation of language guidance.

\begin{table*}[t!]
\centering
\caption{Comparison with ViT-based methods on Office-Home. The best performance is marked as \textbf{bold}.}
\resizebox{0.9\linewidth}{!}{ 
\begin{tabular}{|c|lllllllllllll|}
\hline
Method &A$\to$ C& A$\to$ P & A $\to$ R & C $\to$ A &C $\to$ P &C $\to$ R &P$\to$ A& P$\to$ C & P$ \to$ R & R$ \to$ A &R$ \to$ C &R$ \to$ P& Avg     \\
\hline
\hline
ViT~\cite{DBLP:conf/iclr/DosovitskiyB0WZ21}&54.7& 83.0 &87.2 &77.3 &83.4& 85.5 &74.4 &50.9& 87.2 &79.6& 53.8 &88.8 &75.5\\
CDTrans~\cite{abs-2109-06165}&68.8& 85.0 &86.9& 81.5& 87.1& 87.3 &79.6 &63.3& 88.2 &82.0& 66.0& 90.6& 80.5\\
TVT~\cite{abs-2108-05988}&74.9 &86.8& 89.5& 82.8 &88.0 &88.3 &79.8 &71.9 &90.1 &85.5 &74.6 &90.6 &83.6\\
SSRT~\cite{abs-2204-07683}&75.2 &89.0 &91.1 &85.1 &88.3& 90.0 &85.0 &74.2 &91.3& 85.7& 78.6 &91.8& 85.4\\
DAPL&70.6 &90.2 &91.0& 84.9 &89.2& 90.9 &84.8 &70.5 &90.6 &84.8 &70.1& 90.8 &84.0\\
AD-CLIP&70.9 &92.5 &92.1 &85.4 &92.4 &92.5 &86.7 &74.3 &93.0 &86.9& 72.6& 93.8 &86.1\\
PADCLIP~\cite{lai2023padclip}&76.4& 90.6 &90.8 &86.7 &92.3 &92.0& 86.0& 74.5& 91.5& 86.9 &79.1& 93.1& 86.7\\
PMTrans~\cite{zhu2023patch}&\textbf{81.2}& 91.6 &92.4& \textbf{88.9} &91.6 &93.0 &\textbf{88.5}& 80.0& \textbf{93.4}& \textbf{89.5} &\textbf{82.4}& \textbf{94.5 }&\textbf{88.9}\\
\textbf{CLIP-Div (ViT-B)}&71.4&89.5&89.4&83.0&89.9&89.0&81.5&73.0&89.4&81.8&73.9&90.2&\colorbox{lightgray}{83.5}\\
\textbf{CLIP-Div(ViT-L)} & 79.1&\textbf{ 93.7}& \textbf{93.3}& 87.7& \textbf{94.0}&\textbf{ 93.2}& 86.8 &\textbf{80.6} &\textbf{93.4} &86.8& 80.1& 93.4& \colorbox{lightgray}{88.5}\\
\hline
\end{tabular}}
\label{tab:officehome_vit}
\end{table*}

\begin{table}[t]
\centering
\caption{Comparison with ViT-based methods on Office-31.}
\setlength{\tabcolsep}{1.5mm}
\resizebox{\linewidth}{!}{
\begin{tabular}{|c|lllllll|}
\hline
Method  &A $\to$ W & D$\to$ W & W$\to$ D & A$\to$ D &D$\to$ A &W$\to$ A & Avg \\
\hline
\hline
ViT~\cite{DBLP:conf/iclr/DosovitskiyB0WZ21}&91.2& 99.2& 100. &90.4 &81.1 &80.6& 90.4\\
CDTrans~\cite{abs-2109-06165}&96.7 &99.0& 100.& 97.0& 81.1& 81.9& 92.6\\
SSRT~\cite{abs-2204-07683}&97.7 &99.2& 100. &98.6& 83.5& 82.2 &93.5\\
TVT~\cite{abs-2108-05988}&96.4& 99.4& 100. &96.4 &84.9 &86.1 &93.8\\
PADCLIP~\cite{lai2023padclip}&97.9 &99.2& 100.& 98.5& 84.6 &85.3 &94.3\\
PMTrans~\cite{zhu2023patch}&99.1& 99.6 &100.0 &99.4 &85.7& 86.3& 95.0\\
\textbf{CLIP-Div}&91.8&98.6&99.6&92.8&80.4&81.5&\colorbox{lightgray}{90.8}\\
\textbf{CLIP-Div(ViT-L)}&93.7&98.4&99.5&94.8&\textbf{84.7}&\textbf{85.4}&\colorbox{lightgray}{\textbf{92.9}}\\
\hline
\end{tabular}}
\label{tab:office31_vit}
\end{table}
\begin{table*}[t]
\centering
\caption{Comparison with ViT-based methods on VisDA-2017. The best performance is marked as \textbf{bold}.}
\resizebox{0.995\linewidth}{!}{ 
\begin{tabular}{|c|lllllllllllll|}
\hline
Method &plane& bcycl& bus &car& horse& knife& mcycl& person& plant &sktbrd &train& truck& Avg     \\
\hline
\hline
ViT~\cite{DBLP:conf/iclr/DosovitskiyB0WZ21}& 99.1 &60.7& 70.6& 82.7& 96.5& 73.1 &97.1 &19.7 &64.5& 94.7 &97.2& 15.4 &72.6    \\
CDTrans~\cite{abs-2109-06165}&97.1 &90.5& 82.4& 77.5 &96.6& 96.1 &93.6 &88.6 &97.9 &86.9& 90.3& 62.8 &88.4\\
TVT~\cite{abs-2108-05988}&97.1& 92.9 &85.3 &66.4 &97.1 &97.1 &89.3 &75.5 &95.0 &94.7& 94.5& 55.1 &86.7\\
SSRT~\cite{abs-2204-07683}&98.9& 87.6& 89.1& 84.8 &98.3 &98.7 &96.3 &81.1 &94.9 &97.9& 94.5 &43.1& 88.8\\
DAPL&99.2 &92.5& 93.3 &75.4 &98.6 &92.8 &95.2 &82.5 &89.3 &96.5 &95.1 &63.5& 89.5\\
AD-CLIP&99.6& 92.8& 94.0& 78.6 &98.8& 95.4& 96.8 &83.9& 91.5 &95.8& 95.5 &65.7& 90.7\\
PMTrans~\cite{zhu2023patch}&98.9& 93.7 &84.5 &73.3 &99.0 &98.0 &96.2& 67.8 &94.2 &98.4& 96.6 &49.0 &87.5\\
PADCLIP~\cite{lai2023padclip}&98.1 &93.8 &87.1& 85.5& 98.0 &96.0 &94.4& 86.0 &94.9& 93.3 &93.5 &70.2& 90.9\\
\textbf{CLIP-Div}&\textbf{98.7}&84.5&92.6&77.0&97.2&89.7&\textbf{97.5}&72.0&85.0&\textbf{95.8}&96.1&66.9&\colorbox{lightgray}{87.8}\\
\textbf{CLIP-Div}(ViT-L)&\textbf{98.7}&85.1&\textbf{92.8}&80.2&96.9&88.7&96.8&77.7&86.5&95.6&\textbf{96.8}&\textbf{68.3}&\colorbox{lightgray}{\textbf{88.7}}\\
\hline
\end{tabular}}
\label{tab:VisDA_vit}
\end{table*}

\begin{table*}
\centering
\caption{Comparison with ViT-based methods on DomainNet. The best performance is marked as \textbf{bold}.}
\resizebox{\linewidth}{!}{ 
\begin{tabular}{|c|lllllll||c|lllllll||c|lllllll|}
\hline
ViT~\cite{DBLP:conf/iclr/DosovitskiyB0WZ21}& clp& inf& pnt& qdr& rel& skt& Avg&CDTrans~\cite{abs-2109-06165}& clp& inf& pnt& qdr& rel& skt& Avg&SSRT~\cite{abs-2204-07683}& clp& inf& pnt& qdr& rel& skt& Avg\\
\hline
\hline
clp&- &27.2& 53.1& 13.2& 71.2 &53.3 &43.6&   clp  &  - &29.4& 57.2& 26.0& 72.6& 58.1 &48.7&clp&-& 33.8 &60.2&19.4& 75.8 &59.8 &49.8\\
inf& 51.4& -& 49.3 &4.0 &66.3& 41.1& 42.4&  inf  &   57.0 &- &54.4 &12.8& 69.5& 48.4 &48.4&inf&55.5 &- &54.0 &9.0 &68.2 &44.7& 46.3    \\ 
pnt&53.1 &25.6& - &4.8& 70.0 &41.8 &39.1&  pnt  &  62.9& 27.4& -& 15.8& 72.1& 53.9& 46.4&pnt&61.7& 28.5& -& 8.4& 71.4& 55.2& 45.0   \\
qdr&30.5& 4.5 &16.0 &- &27.0 &19.3& 19.5 & qdr &  44.6& 8.9 &29.0 &- &42.6 &28.5 &30.7&qdr&42.5 &8.8 &24.2& - &37.6 &33.6 &29.3      \\
rel&58.4 &29.0 &60.0 &6.0& - &45.8 &39.9  &rel &  66.2& 31.0& 61.5& 16.2& -& 52.9& 45.6&rel&69.9 &37.1& 66.0& 10.1& - &58.9& 48.4     \\
skt&63.9 &23.8& 52.3 &14.4 &67.4& -& 44.4 & skt  &  69.0& 29.6 &59.0& 27.2 &72.5& -& 51.5&skt&70.6 &32.8 &62.2 &21.7 &73.2 &- &52.1     \\ 
Avg&51.5 &22.0 &46.1 &8.5& 60.4 &40.3 &\colorbox{lightgray}{38.1}&   Avg  & 59.9 &25.3 &52.2& 19.6& 65.9 &48.4& \colorbox{lightgray}{45.2}&Avg&60.0& 28.2& 53.3& 13.7& 65.3 &50.4& \colorbox{lightgray}{45.2}\\
\hline
\hline
PMTrans~\cite{zhu2023patch} & clp& inf& pnt& qdr& rel& skt& Avg&PADCLIP~\cite{lai2023padclip} & clp& inf& pnt& qdr& rel& skt& Avg&CLIP-Div& clp& inf& pnt& qdr& rel& skt& Avg \\
\hline
\hline
clp & - & 34.2 &62.7 &32.5 &79.3 &63.7 & 54.5  & clp& -& 55.1& 71.1 &36.8 &84.2& 68.1 &61.3&clp&-&39.4&66.2&23.8&80.4&64.4&54.8\\
inf& 67.4&- &61.1 & 22.2 & 78.0 & 57.6&57.3 & inf &73.6& - &70.6 &18.0 &83.5 &66.6 &62.5&inf&73.9&-&65.7&13.6&80.1&63.8&59.4 \\ 
pnt&69.7&33.5&-&23.9&79.8&61.2&53.6 &pnt& 75.4& 54.3& - &32.0 &83.5 &67.2& 62.5&pnt&74.5&39.6&-&16.6&81.4&65.0&55.4 \\
qdr&54.6&17.4&38.9&-&49.5  &41.0&40.3  &qdr& 74.6& 53.6& 70.0&- &83.1& 66.1 &69.5&qdr&70.9&15.1&38.2&-&74.2&51.9&50.1\\
rel&74.1&35.3&70.0&25.4&-&61.1&53.2 & rel &76.4 &54.9& 72.7 &31.7 &- &67.5 &60.6&rel&74.3&39.3&66.3&18.2&-&63.4&52.3\\
skt&73.8&33.0&62.6&30.9&77.5&-& 55.6 & skt& 76.3 &54.9 &71.7& 34.9 &83.6& -& 64.3&skt&75.6&39.0&67.3&19.0&78.3&-&55.8\\ 
Avg&67.9&30.7&59.1&27.0&72.8&56.9&\colorbox{lightgray}{\textbf{52.4}}  &Avg& 75.3 &54.6& 71.2& 30.7& 83.6 &67.1& \colorbox{lightgray}{63.7}&Avg&73.8&34.5&60.7&18.2&78.9&61.7&\colorbox{lightgray}{\textbf{54.6}}\\
\hline
\end{tabular}}
\label{tab:domainnet_vit}
\end{table*}




\clearpage

\subsection{Distinctions with Knowledge Distillation}

In this work, our intuition is that: \textit{compared with directly aligning source and target domains, reducing the gap between the two distributions and the domain-agnostic distribution -- as a crucial bridge -- can facilitate the domain alignment}. Therefore, we propose two different prompts and divergence measurement losses to measure the domain divergence between the domain-specific distributions and the domain-agnostic distribution. Subsequently, we align the source and target domains via minimizing the loss terms. Different from the traditional KD methods, we utilize cosine similarity and KL divergence to measure the divergence between the domain-specific distributions and the domain-agnostic distribution instead of directly transferring knowledge from CLIP to the UDA model. 

\subsection{Functions Used in $\mathcal{L}_{abs}$ and $\mathcal{L}_{rel}$}

Due that $\mathcal{G}(\mathcal{F}(\boldsymbol{x}^{s})$ is the prediction probability, we utilize the KL divergence to measure the divergence. And we use cosine similarity over KL divergence for $\mathcal{L}_{rel}$ as $\Delta_1$ and $\Delta_2$ are distances not probability distributions.

\begin{figure*}[t]
    \centering
    \includegraphics[width=\linewidth]{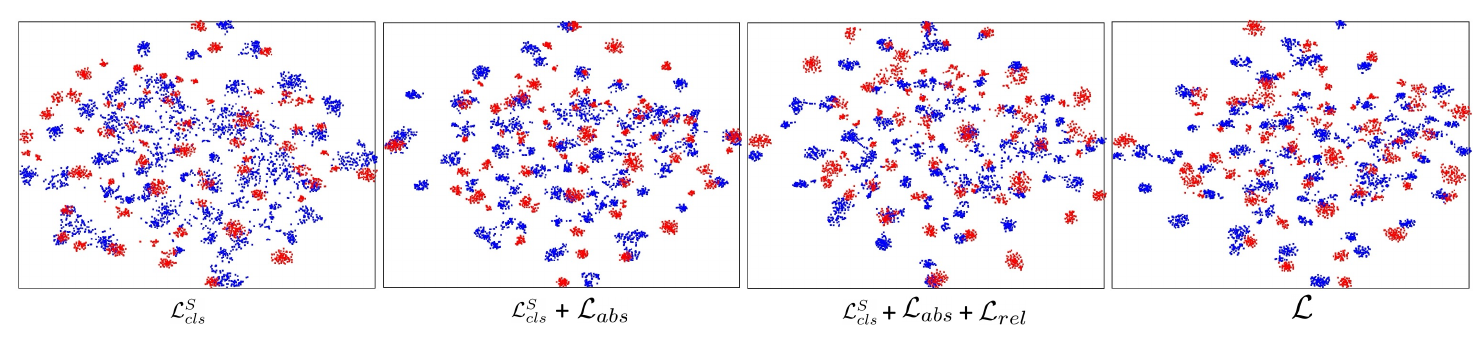}
    \caption{The t-SNE visualization with the proposed losses on the task A $\to$ P.}
    \label{fig:t-sne1}
\end{figure*}

\subsection{Effect of CLIP for Domain Adaptation or Pseudo-Labeling}

Given our proposal to designate prompts and fix CLIP to derive the domain-agnostic distribution, it is possible that the domain-agnostic distribution may not be optimal compared to utilizing prompt learning or fine-tuning CLIP. Nevertheless, the outcomes of comparative analysis and ablation study consistently validate the effectiveness of our proposed language-guided approach for UDA. Moreover, t-SNE~\cite{van2008visualizing} visualizations in Fig. \ref{fig:t-sne1} offer additional evidence of its effectiveness with language guidance.

\clearpage

\section{Applications on Source-Free Domain Adaptation}
\label{source-free}
To evaluate the generalization capability of CLIP-Div, we leverage its application in addressing source-free domain adaptation (SFDA)  challenges. Our experimentation focuses on the Office-Home dataset, where we compare CLIP-Div against SoTA SFDA methodologies, including G-SFDA~\cite{yang2021generalized}, SHOT~\cite{liang2020we}, NRC~\cite{yang2021exploiting}, A$^2$Net~\cite{xia2021adaptive}, and SFDA-DE~\cite{ding2022source}. The results, as presented in Tab.~\ref{tab:officehome_sourcefree}, unequivocally demonstrate that CLIP-Div outperforms all considered SFDA methods across various transfer tasks. Notably, CLIP-Div attains an impressive \textbf{82.3}\% average accuracy, showcasing a remarkable \textbf{9.4}\% accuracy improvement. These findings underscore the robust generalization capacity of our proposed approach in diverse UDA scenarios and affirm the effectiveness of CLIP-Div.

\begin{figure*}[t]
    \centering
    \includegraphics[width=0.7\linewidth]{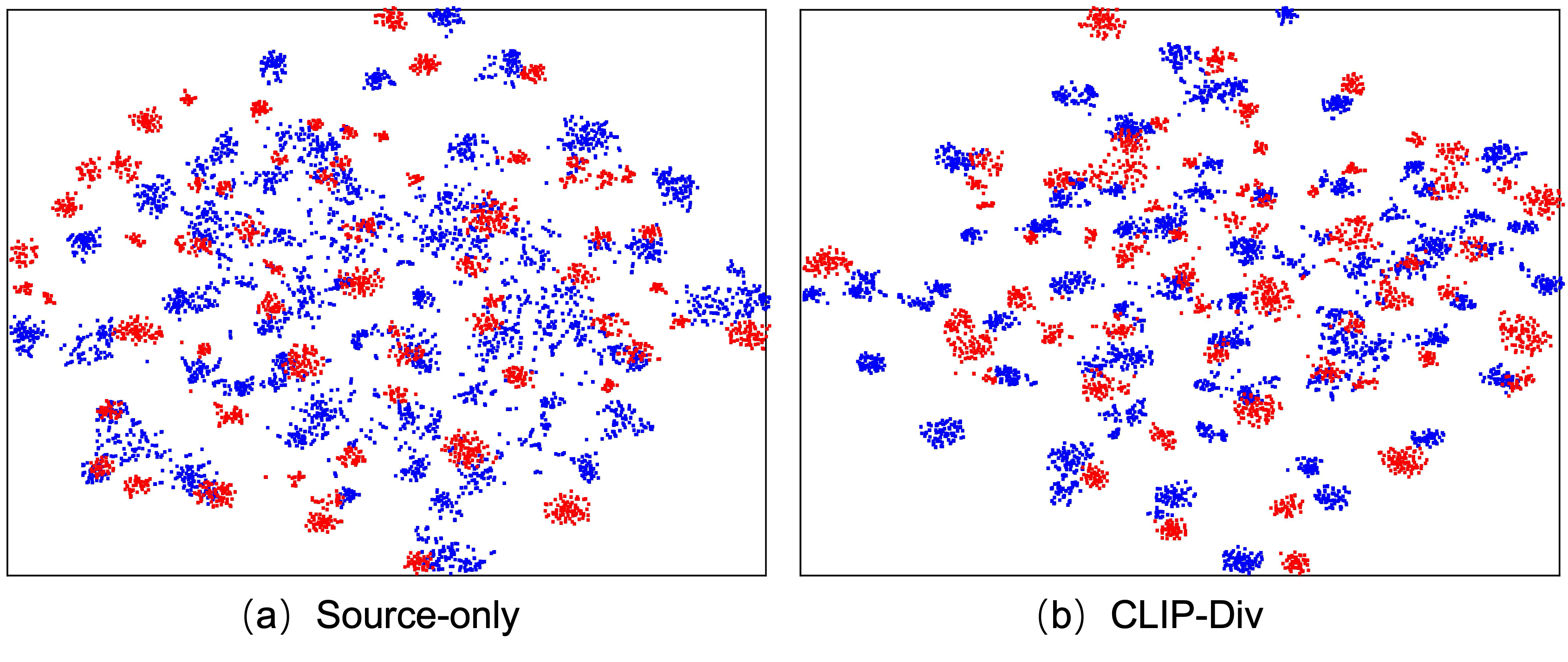}
    \caption{The visualization of embedded features on the
task A $\to$ P on Office-Home. Red and blue points denote the source and
target domains, respectively.}
    \label{fig:tsne}
\end{figure*}

\begin{table*}[t!]
\centering
\caption{Comparison with SoTA SFDA methods on Office-Home. The best performance is marked as \textbf{bold}.}
\resizebox{0.9\linewidth}{!}{ 
\begin{tabular}{|c|lllllllllllll|}
\hline
Method &A$\to$ C& A$\to$ P & A $\to$ R & C $\to$ A &C $\to$ P &C $\to$ R &P$\to$ A& P$\to$ C & P$ \to$ R & R$ \to$ A &R$ \to$ C &R$ \to$ P& Avg     \\
\hline
\hline
ResNet-50~\cite{he2016deep} & 44.9& 66.3 &74.3 &51.8& 61.9 &63.6& 52.4 &39.1& 71.2& 63.8& 45.9& 77.2 &\colorbox{lightgray}{59.4}   \\
G-SFDA~\cite{yang2021generalized}&57.9 &78.6 &81.0& 66.7 &77.2& 77.2& 65.6& 56.0& 82.2& 72.0 &57.8 &83.4& \colorbox{lightgray}{71.3}   \\
SHOT~\cite{liang2020we}&57.1 &78.1 &81.5& 68.0 &78.2& 78.1& 67.4& 54.9& 82.2& 73.3 &58.8 &84.3& \colorbox{lightgray}{71.8}   \\
NRC~\cite{yang2021exploiting}&57.7 &80.3 &82.0& 68.1 &79.8& 78.6& 65.3& 56.4& 83.0& 71.0 &58.6 &85.6&\colorbox{lightgray} {72.2}   \\
$A^{2}$Net~\cite{xia2021adaptive}&58.4 &79.0 &82.4& 67.5 &79.3& 78.9& 68.0& 56.2& 82.9& 74.1 &60.5 &85.0& \colorbox{lightgray}{72.8}   \\
SFDA-DE~\cite{ding2022source}&59.7 &79.5 &82.4& 69.7 &78.6& 79.2& 66.1& 57.2& 82.6& 73.9 &60.8 &85.5& \colorbox{lightgray}{72.9}   \\
\textbf{CLIP-Div}&\textbf{68.9}&\textbf{88.6}&\textbf{88.3}&\textbf{81.5}&\textbf{88.7}&\textbf{88.6}&\textbf{80.7}&\textbf{72.0}&\textbf{88.4}&\textbf{81.4}&\textbf{71.5}&\textbf{89.3}&\colorbox{lightgray}{\textbf{82.3}}\\
\hline
\end{tabular}}
\label{tab:officehome_sourcefree}
\end{table*}

\section{Visual Representation}
\label{visualization}

Fig.~\ref{fig:tsne} provides a visual representation of the embedded features in the task A $\to$ P using t-SNE~\cite{van2008visualizing}. In the case of utilizing only the source data, the embedded target domain features are positioned around the clusters of the source domain features, yet they do not coalesce into distinct clusters themselves. In contrast, our proposed method, CLIP-Div, effectively establishes compact clusters of target domain features in proximity to the source domain features. This observation substantiates the effectiveness of our approach in addressing UDA tasks with language guidance.

\clearpage

\section{Algorithm}
\label{algorithm}
The overall algorithm of CLIP-Div is shown in Algorithm.\ref{alg}.
\begin{algorithm}[h]
	\caption{The Proposed CLIP-Div} 
	\label{alg} 
	\begin{algorithmic}[1]
	    \STATE \textbf{Input}: $\mathcal{D}^s=\left\{\left(x_i^s, y_i^s\right)\right\}_{i=1}^{n_s}$, $\mathcal{D}^t=\left\{\left(x_i^t\right)\right\}_{i=1}^{n_t}$, $\boldsymbol{prompt}^{s}$, $\boldsymbol{prompt}^{t}$, $\boldsymbol{prompt}^{a}$; 
            \\ \textbf{Max epochs}: T
	    \\ \textbf{Model}:  $\mathcal{F}$, $\mathcal{G}$, $\mathcal{E_I}$, $\mathcal{E_T}$;
	    \FOR{for t $\xleftarrow[]{}$ 1 to $T_1$}
    	    \STATE Compute classification loss for the source data:
                     $  \mathcal{L}_{cls}^{s}=\mathbb{E}_{\left(\boldsymbol{x}^{s}, \hat{\boldsymbol{y}}^{s}\right) \sim D^{s}} \ell\left(\mathcal{G}\left(\mathcal{F}\left(\boldsymbol{x}^{s}\right)\right), \boldsymbol{y}^{s}\right)$ ;
                \STATE Compute the absolute divergence loss for the source and target data:\\
                $\mathcal{L}_{abs}^{s} = \mathbb{E}_{\boldsymbol{x}^{s} \sim D^{s}} KL(\mathcal{G}(\mathcal{F}(\boldsymbol{x}^{s}))\|{p}(\boldsymbol{x}^{s}, \boldsymbol{prompt}^{a}))$,\\
    $\mathcal{L}_{abs}^{t} = \mathbb{E}_{\boldsymbol{x}^{t} \sim D^{t}} KL( \mathcal{G}(\mathcal{F}(\boldsymbol{x}^{t}))\| {p}(\boldsymbol{x}^{t}, \boldsymbol{prompt}^{a}));$\\
               \STATE Compute the total absolute divergence loss:\\
               $ \mathcal{L}_{abs} = \mathcal{L}_{abs}^{s}+\mathcal{L}_{abs}^{t}$;
               \STATE Obtain the domain-agnostic text representations:\\
             $  \mathcal{E_{T}}({\boldsymbol{prompt}}^{avg}_{k}) = \frac{\mathcal{E_{T}}({\boldsymbol{prompt}}_{k}^s)+\mathcal{E_{T}}({\boldsymbol{prompt}}_{k}^t)}{2},$\\
              ${p}_k(\boldsymbol{x}, \boldsymbol{prompt}^{avg})=\frac{\exp \left(\cos \left(\mathcal{E_{T}}({\boldsymbol{prompt}}^{avg}_{k}), \mathcal{E_{I}}\left(\boldsymbol{x}\right) / \tau \right)\right.}{\sum_{k=1}^K \exp \left(\cos \left(\mathcal{E_{T}}({\boldsymbol{prompt}}^{avg}_{k}), \mathcal{E_{I}}\left(\boldsymbol{x}\right) / \tau \right)\right.}; $ 
               \STATE Compute the relative divergence loss:\\
                $\Delta_{1}={p}(\boldsymbol{x}^{s}, \boldsymbol{prompt}^{avg})-{p}(\boldsymbol{x}^{t}, \boldsymbol{prompt}^{avg}), $\\
    $\Delta_{2}=\mathcal{G}(\mathcal{F}(\boldsymbol{x}^{s}))-\mathcal{G}(\mathcal{F}(\boldsymbol{x}^{t}))$,\\
    $\mathcal{L}_{rel}=1-\frac{\Delta_1\cdot \Delta_{2}}{|\Delta_1||\Delta_2|}$;
    \STATE Obtain the pseudo labels for target data:

  $  c_k^t=\frac{\sum_{\boldsymbol{x}^{t} \sim D^{t}} (\delta_k^t+{p}_k^t(\boldsymbol{x}^t, \boldsymbol{prompt}^t)) \mathcal{F}\left(\boldsymbol{x}^{t}\right)}{\sum_{\boldsymbol{x}^{t} \sim D^{t}} (\delta_k^t+{p}_k^t(\boldsymbol{x}^t,\boldsymbol{prompt}^t))},$\\
 $  \hat{y}^t=\arg \min _k d\left(\mathcal{F}\left(\boldsymbol{x}^{t}\right), c_k^t\right),$\\
 ${c_k^t}^{\prime}  =\frac{\sum_{\boldsymbol{x}^{t} \sim D^{t}} \mathbb{I}\left(\hat{y}^t=k\right) \mathcal{F}\left(\boldsymbol{x}^t\right)}{\sum_{\boldsymbol{x}^{t} \sim D^{t}} \mathbb{I}\left(\hat{y}^t=k\right)}$, \\
${\hat{y}^t}  =\arg \min _k d\left(\mathcal{F}\left(\boldsymbol{x}^{t}\right), {c_k^t}^{\prime} \right)$,\\

\STATE Calculate the classification loss for target data based on pseudo labels:

$\mathcal{L}_{cls}^{t}=\mathbb{E}_{\left(\boldsymbol{x}^{t}, \hat{\boldsymbol{y}}^{t}\right) \sim D^{t}} \ell\left(\mathcal{G}\left(\mathcal{F}\left(\boldsymbol{x}^{t}\right)\right), \hat{\boldsymbol{y}}^{t}\right);$

\STATE Calculate the total loss:

$\mathcal{L}= \mathcal{L}_{cls}^{s}+\lambda_{abs}*\mathcal{L}_{abs}+\lambda_{rel}*\mathcal{L}_{rel}+\lambda_{pl}*\mathcal{L}_{pl};$

\STATE Back propagation for $\mathcal{L}$

         \STATE Update the UDA model $\mathcal{G}(\mathcal{F}(\cdot))$.
    	 \ENDFOR
	    \STATE  \textbf{return}  Weights of the UDA model $\mathcal{G}(\mathcal{F}(\cdot))$.
	    \STATE  \textbf{End}.
	\end{algorithmic} 
\end{algorithm}

\clearpage
\section{Prompt Examples}
\label{prompt_example}

We present prompt examples on the Office-Home dataset to enhance comprehension of our methodology. Specifically, domain-specific prompt examples related to the Real-World and Clipart domains, and domain-agnostic prompt examples are provided as below.

\noindent\textbf{Domain-specific prompts in Real-World:}

\begin{lstlisting}[language=Python]
a real world photo of a Alarm Clock
a real world photo of a Backpack
a real world photo of a Batteries
a real world photo of a Bed
a real world photo of a Bike
a real world photo of a Bottle
a real world photo of a Bucket
a real world photo of a Calculator
a real world photo of a Calendar
a real world photo of a Candles
a real world photo of a Chair
a real world photo of a Clipboards
a real world photo of a Computer
a real world photo of a Couch
a real world photo of a Curtains
a real world photo of a Desk Lamp
a real world photo of a Drill
a real world photo of a Eraser
a real world photo of a Exit Sign
a real world photo of a Fan
a real world photo of a File Cabinet
a real world photo of a Flipflops
a real world photo of a Flowers
a real world photo of a Folder
a real world photo of a Fork
a real world photo of a Glasses
a real world photo of a Hammer
a real world photo of a Helmet
a real world photo of a Kettle
a real world photo of a Keyboard
a real world photo of a Knives
a real world photo of a Lamp Shade
a real world photo of a Laptop
a real world photo of a Marker
a real world photo of a Monitor
a real world photo of a Mop
a real world photo of a Mouse
a real world photo of a Mug
a real world photo of a Notebook
a real world photo of a Oven
a real world photo of a Pan
a real world photo of a Paper Clip
a real world photo of a Pen
a real world photo of a Pencil
a real world photo of a Postit Notes
a real world photo of a Printer
a real world photo of a Push Pin
a real world photo of a Radio
a real world photo of a Refrigerator
a real world photo of a Ruler
a real world photo of a Scissors
a real world photo of a Screwdriver
a real world photo of a Shelf
a real world photo of a Sink
a real world photo of a Sneakers
a real world photo of a Soda
a real world photo of a Speaker
a real world photo of a Spoon
a real world photo of a TV
a real world photo of a Table
a real world photo of a Telephone
a real world photo of a ToothBrush
a real world photo of a Toys
a real world photo of a Trash Can
a real world photo of a Webcam
\end{lstlisting}

\noindent\textbf{Domain-specific prompts in Clipart:}

\begin{lstlisting}[language=Python]
a clipart photo of a Alarm Clock
a clipart photo of a Backpack
a clipart photo of a Batteries
a clipart photo of a Bed
a clipart photo of a Bike
a clipart photo of a Bottle
a clipart photo of a Bucket
a clipart photo of a Calculator
a clipart photo of a Calendar
a clipart photo of a Candles
a clipart photo of a Chair
a clipart photo of a Clipboards
a clipart photo of a Computer
a clipart photo of a Couch
a clipart photo of a Curtains
a clipart photo of a Desk Lamp
a clipart photo of a Drill
a clipart photo of a Eraser
a clipart photo of a Exit Sign
a clipart photo of a Fan
a clipart photo of a File Cabinet
a clipart photo of a Flipflops
a clipart photo of a Flowers
a clipart photo of a Folder
a clipart photo of a Fork
a clipart photo of a Glasses
a clipart photo of a Hammer
a clipart photo of a Helmet
a clipart photo of a Kettle
a clipart photo of a Keyboard
a clipart photo of a Knives
a clipart photo of a Lamp Shade
a clipart photo of a Laptop
a clipart photo of a Marker
a clipart photo of a Monitor
a clipart photo of a Mop
a clipart photo of a Mouse
a clipart photo of a Mug
a clipart photo of a Notebook
a clipart photo of a Oven
a clipart photo of a Pan
a clipart photo of a Paper Clip
a clipart photo of a Pen
a clipart photo of a Pencil
a clipart photo of a Postit Notes
a clipart photo of a Printer
a clipart photo of a Push Pin
a clipart photo of a Radio
a clipart photo of a Refrigerator
a clipart photo of a Ruler
a clipart photo of a Scissors
a clipart photo of a Screwdriver
a clipart photo of a Shelf
a clipart photo of a Sink
a clipart photo of a Sneakers
a clipart photo of a Soda
a clipart photo of a Speaker
a clipart photo of a Spoon
a clipart photo of a TV
a clipart photo of a Table
a clipart photo of a Telephone
a clipart photo of a ToothBrush
a clipart photo of a Toys
a clipart photo of a Trash Can
a clipart photo of a Webcam
\end{lstlisting}

\noindent\textbf{Domain-agnostic prompts:}

\begin{lstlisting}[language=Python]
a photo of a Alarm Clock
a photo of a Backpack
a photo of a Batteries
a photo of a Bed
a photo of a Bike
a photo of a Bottle
a photo of a Bucket
a photo of a Calculator
a photo of a Calendar
a photo of a Candles
a photo of a Chair
a photo of a Clipboards
a photo of a Computer
a photo of a Couch
a photo of a Curtains
a photo of a Desk Lamp
a photo of a Drill
a photo of a Eraser
a photo of a Exit Sign
a photo of a Fan
a photo of a File Cabinet
a photo of a Flipflops
a photo of a Flowers
a photo of a Folder
a photo of a Fork
a photo of a Glasses
a photo of a Hammer
a photo of a Helmet
a photo of a Kettle
a photo of a Keyboard
a photo of a Knives
a photo of a Lamp Shade
a photo of a Laptop
a photo of a Marker
a photo of a Monitor
a photo of a Mop
a photo of a Mouse
a photo of a Mug
a photo of a Notebook
a photo of a Oven
a photo of a Pan
a photo of a Paper Clip
a photo of a Pen
a photo of a Pencil
a photo of a Postit Notes
a photo of a Printer
a photo of a Push Pin
a photo of a Radio
a photo of a Refrigerator
a photo of a Ruler
a photo of a Scissors
a photo of a Screwdriver
a photo of a Shelf
a photo of a Sink
a photo of a Sneakers
a photo of a Soda
a photo of a Speaker
a photo of a Spoon
a photo of a TV
a photo of a Table
a photo of a Telephone
a photo of a ToothBrush
a photo of a Toys
a photo of a Trash Can
a photo of a Webcam

\end{lstlisting}

\clearpage

\bibliographystyle{./IEEEtran}

\bibliography{./IEEEabrv,./egbib}